\documentclass[10pt]{article}
\usepackage[T1]{fontenc}
\usepackage{newpxtext}
\usepackage[scaled=0.92]{helvet}

\usepackage{tim_preprint}

\usepackage{hyperref}
\definecolor{citeblue}{HTML}{2E6AB1}\definecolor{linkblue}{HTML}{1F5A96}
\hypersetup{colorlinks=true,linkcolor=linkblue,citecolor=citeblue,urlcolor=linkblue}
\usepackage{url}
\usepackage{xspace}

\title{On the Origin and Evolution of Training-Inference Mismatch in LLM Reinforcement Learning}
\shorttitle{Origin and Evolution of Training-Inference Mismatch}
\preprintline{Preprint, October 2026}
\papercode{\href{https://github.com/Skylanding/Training-Inference-Mismatch}{\texttt{https://github.com/Skylanding/Training-Inference-Mismatch}}}
\author{%
    Yu Li\textsuperscript{1,*}\quad
    Yufa Zhou\textsuperscript{2,*}\quad
    Tian Lan\textsuperscript{1}
}
\affiliations{%
    \textsuperscript{1}George Washington University\quad
    \textsuperscript{2}Duke University
}
\date{}

\usepackage{amsmath,amsthm}
\usepackage{newpxmath}

\usepackage{amsmath,amsfonts,bm}

\def\eqref#1{equation~\ref{#1}}

\def\1{\bm{1}}

\DeclareMathAlphabet{\mathsfit}{\encodingdefault}{\sfdefault}{m}{sl}
\SetMathAlphabet{\mathsfit}{bold}{\encodingdefault}{\sfdefault}{bx}{n}

\usepackage{xcolor}
\usepackage{graphicx}
\usepackage{booktabs,multirow}
\usepackage{colortbl}
\definecolor{mPG}{HTML}{8C8C8C}\definecolor{mPPO}{HTML}{C9A055}\definecolor{mSC}{HTML}{2F5D8C}
\definecolor{mTIS}{HTML}{7A5BA8}\definecolor{mTISSC}{HTML}{D9773A}\definecolor{mSLC}{HTML}{1B7F5A}

\newcommand{\grouprow}[2]{\rowcolor{black!6}\multicolumn{#1}{@{\hspace{4pt}}l}{\footnotesize\itshape\color{black!72}#2}}
\usepackage{algorithm}
\usepackage[noend]{algpseudocode}
\algrenewcommand\algorithmicrequire{\textsf{Input:}}
\newcommand{\tabhead}[1]{#1}
\usepackage{float}
\usepackage{pifont}
\usepackage[capitalise,noabbrev]{cleveref}
\usepackage{enumitem}
\newcommand{\cmark}{\textcolor{mSLC}{\ding{51}}}
\newcommand{\xmark}{\textcolor{black!35}{\ding{55}}}
\definecolor{oursbg}{HTML}{E4EDF8}\definecolor{oursfg}{HTML}{1F4E9C}

\newcommand{\cn}[1]{\ding{\numexpr181+#1\relax}}
\newcommand{\old}{\theta_\mathrm{old}}
\usepackage{tabularx}
\usepackage[bottom]{footmisc}
\newcommand{\Ex}{\mathbb{E}}

\begin{document}

\maketitle
{\let\thefootnote\relax\footnotetext{%
    $^*$Equal contribution.
    Correspondence: \href{mailto:yul@gwu.edu}{\texttt{yul@gwu.edu}},
    \href{mailto:yufa.zhou@duke.edu}{\texttt{yufa.zhou@duke.edu}},
    \href{mailto:tlan@gwu.edu}{\texttt{tlan@gwu.edu}}.}}

\begin{abstract}

On-policy reinforcement learning for large language models assumes that training responses are sampled from the current policy.
In practice, an inference engine samples the responses and a separate trainer computes the gradients, and the \textbf{training-inference mismatch} between the two makes the training data off-policy. The mismatch has two sources, the numerics of the engine and the stale weights of asynchronous sampling, and a correction that stabilizes training under one source can fail under the other. \textit{We ask which property of a mismatch source decides the correction it needs.} Evaluating the trainer at the sampler's weights splits each sampled-token error exactly into a \textit{numerical discrepancy} of the engine and a \textit{policy lag} of the trainer, and we follow both through training. The numerical discrepancy stays approximately constant, so it biases the updates by a nearly fixed amount, and centering the score at each prefix removes its leading bias. The policy lag is produced by the trainer's own updates: the bias it leaves enters the next update, which enlarges the next lag and closes a feedback loop. Importance weighting cuts this loop only while the lag ratio stays concentrated, and on Qwen3-1.7B it fails at the step where the effective sample size of the ratio falls below one half. From this account we derive the \textbf{discrepancy-lag correction} (DLC), which centers the numerical discrepancy, clips the lag ratio against the sampler's weights and resynchronizes the sampler when this effective sample size falls below one half. Under a stale sampler, DLC keeps the KL divergence from on-policy training within twice the KL between two independent on-policy trainings on Qwen3-0.6B and Qwen3-1.7B, and its resynchronization prevents collapses that the per-token corrections leave and that a KL measured on the sampler's prefixes does not detect.

\end{abstract}

\begin{figure}[t]
\centering
\sbox0{\begin{minipage}[b]{4.07in}\centering
{\small\bfseries (a) Origin, evolution and correction of the mismatch}\\[3pt]
\includegraphics[height=3.05in]{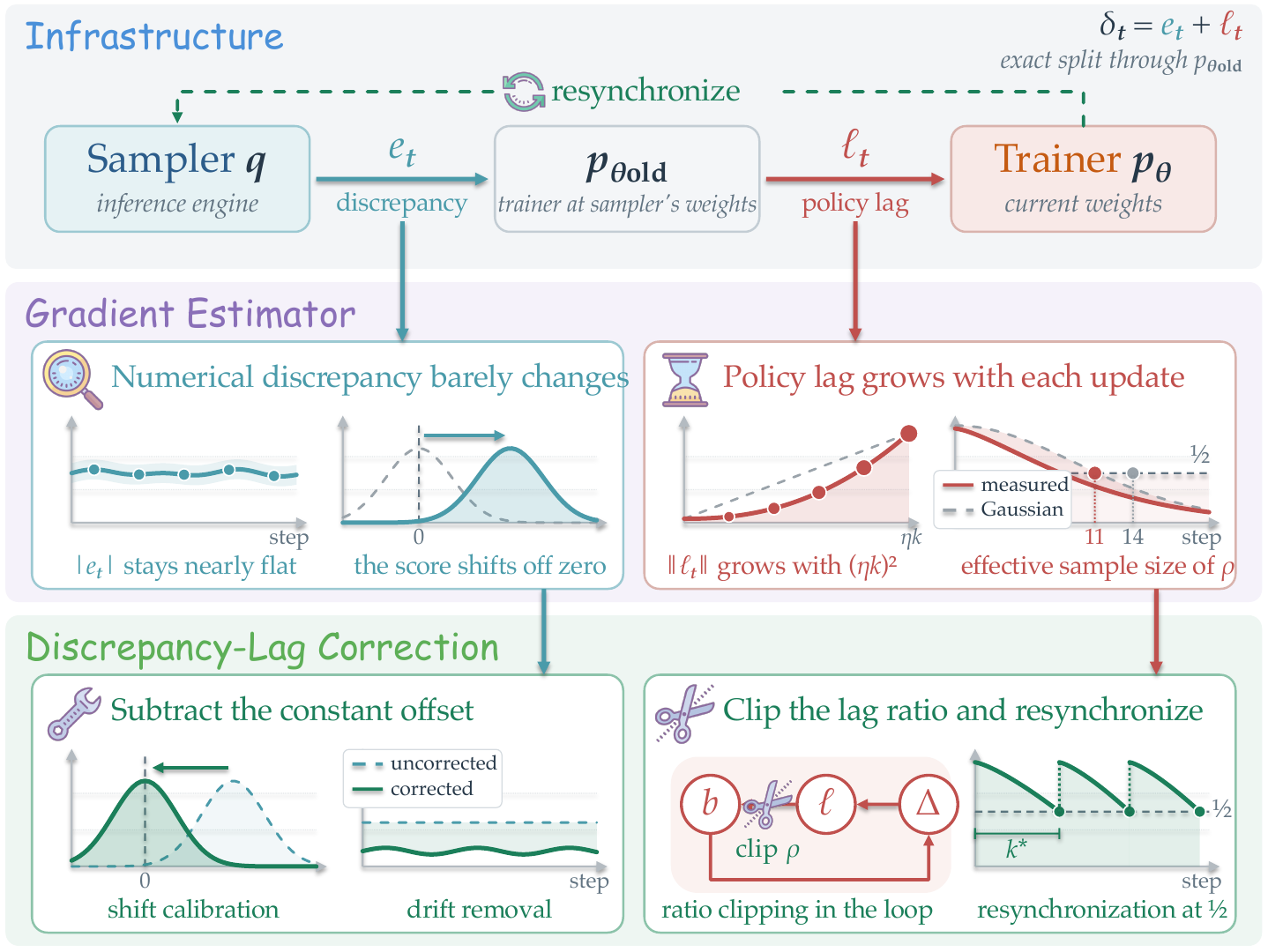}
\end{minipage}}%
\usebox0\hfill\includegraphics[height=\dimexpr\ht0+\dp0\relax]{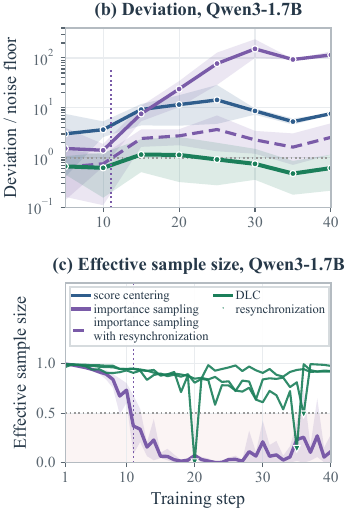}
\caption{\textbf{Origin of the training-inference mismatch and the correction built on it.} The trainer's forward pass at the sampler's weights splits the mismatch exactly into the numerical discrepancy and the policy lag, and the gradient estimator receives a constant offset from the first and a growing lag ratio from the second. DLC subtracts the offset, clips the lag ratio inside the loop through the update and resynchronizes the sampler when the effective sample size of the lag ratio falls below one half \textbf{(a)}. Under a stale sampler, DLC keeps Qwen3-1.7B at the noise floor through step 40, while importance sampling leaves the floor once its effective sample size falls below one half at step 11 \textbf{(b)}. The three DLC trainings resynchronize the sampler when the effective sample size of their lag ratio falls below one half, and the effective sample size of importance sampling stays below one half after step 11 \textbf{(c)}.}
\label{fig:teaser}
\end{figure}

\section{Introduction}

Reinforcement learning for language models \citep{shao2024deepseekmath,li2026arise,zhou2026athanor} typically runs two systems on the same weights. Inference engines such as vLLM \citep{kwon2023vllm,zheng2024sglang,li2026momentkv} sample the responses, and a trainer computes the policy gradient on them. The engine computes token probabilities with its own kernels and precision, so its distribution $q$ differs from the trainer's distribution $p$ even at identical weights \citep{yao2025offpolicy,liu2025speed}, and low-precision samplers widen the gap. Asynchronous systems such as AReaL \citep{fu2025areal} and PipelineRL \citep{piche2025pipelinerl} add a second source, since the sampler keeps the weights of the last synchronization while the trainer moves on. Training on samples from $q$ with gradients of $p$ is therefore off-policy. We call the whole gap between $q$ and $p$, from both sources, the \textit{training-inference mismatch}. Prior work addresses this mismatch with three corrections \citep{yao2025offpolicy,marek2026score,schulman2017ppo}: (1) truncated importance sampling weights each sampled token by the truncated ratio $p/q$; (2) score centering subtracts the sampler's mean score at each prefix and thereby removes the drift, the part of the update that depends on the advantages only through their mean; (3) PPO clips the ratio of the trainer to a reference policy at each token.

The outcome of these corrections depends on the source of the mismatch. Score centering keeps training stable with a quantized sampler and falls behind importance sampling with a stale one, while PPO trains under staleness and fails under quantization \citep{marek2026score}. Under staleness alone the collapse of asynchronous GRPO scales with the product of staleness and learning rate \citep{song2026scaling}, and importance-weighted asynchronous training loses its effective sample size before it diverges \citep{Huang2026StableAV}. The behavior of a correction therefore depends jointly on the estimator and on the process that produces the mismatch. At a single prefix, however, the untruncated ratio and centering both remove the drift, so the source must act through what each correction leaves after the drift and through how that remainder evolves during training. The question is which property of a mismatch source decides the correction it needs.

We trace the mismatch to its origin in the two layers of the system and follow its evolution during training. The infrastructure decides how far the sampler departs from the trainer, and the algorithm corrects what reaches the gradient. Once the infrastructure aligns sampling, reporting and token sequences, the mismatch reaches the algorithm as a numerical discrepancy and a policy lag, and the trainer's log-probability at the sampler's weights splits the sampled-token error exactly into these two terms, as \cref{fig:teaser} illustrates. The numerical discrepancy originates in the engine and stays approximately fixed during training. The policy lag originates in the trainer's own updates, the residual bias it leaves enters the next update, and the next update sets the next lag. Corrections of the policy lag therefore have to keep this loop from amplifying the update. Our contributions follow this account.

\begin{itemize}[leftmargin=1.6em]
\item[\cn{1}] Following the factorization of the importance ratio at the sampler's weights \citep{guan2026missing}, we split the sampled-token error exactly into the numerical discrepancy and the policy lag. After centering removes the drift, the remaining bias consists of a prefix, a continuation, a token and a tail term, which keep a fixed size for the numerical discrepancy and grow with the accumulated update for the policy lag.

\item[\cn{2}] For the policy lag, the token and prefix terms enter the next update and close a loop, which locates the part of the known staleness bias \citep{song2026scaling,Xu2026GACSA} that survives centering. As a consequence, the excess update norm of score centering over importance sampling has a positive quadratic term in the product $\eta k$ of learning rate and step count. Ratio weighting removes the token term within its validity range, which ends where the effective sample size of the lag ratio falls below one half. On Qwen3-1.7B the heavy tail of the ratio brings this crossing to step 11, three steps before the Gaussian prediction, and importance sampling leaves the noise floor at that step.

\item[\cn{3}] The discrepancy-lag correction (DLC) centers the numerical discrepancy, clips the lag ratio against the sampler's weights \citep{chang2026calibrating} and resynchronizes the sampler at the end of the validity range. Under a stale sampler DLC keeps the deviation from on-policy training within twice the noise floor on Qwen3-0.6B and Qwen3-1.7B, and on Qwen3-1.7B the resynchronization prevents the collapses that the clip alone leaves.
\end{itemize}

\section{Background}\label{sec:background}

\subsection{Training-inference mismatch}
\Cref{tab:directions} lists the mismatch by the layer that corrects it. Once the infrastructure aligns sampling, reporting and token sequences, only the weight version and the engine numerics reach the algorithm, and only the weight version changes with the update.
The other three rows of the infrastructure layer concern what the sampler reports and what the trainer receives. Top-p sampling and processed log-probabilities make the reported $q$ differ from the distribution that drew the token, so the sampler has to report the drawn distribution. Retokenization of the decoded text can change the token sequence, and passing token ids to the trainer keeps the sequence identical on both sides. Filtering responses by their reward and deferring long responses in partial rollouts make the composition of the batch depend on the outcome, and outcome-independent batching keeps the batch a sample of $q$. These three mismatches vanish under the configuration of \cref{sec:setup}, so the weight version and the engine numerics remain as the two sources that the algorithm has to correct. In the algorithm layer, the weight version gives the policy lag, which grows with the update, and decoupled PPO and M2PO weight the ratio of the trainer to the sampler's weights, while VCPO scales the learning rate by the effective sample size. The engine numerics give the numerical discrepancy, which stays approximately fixed in training, and truncated importance sampling and score centering correct it at each sampled token.

\begin{table}[!htb]
\centering
\caption{\textbf{Training-inference mismatch and its corrections in the two layers.}}
\label{tab:directions}
\small
\renewcommand{\arraystretch}{1.2}
\begin{tabularx}{\linewidth}{@{}p{0.21\linewidth} >{\raggedright\arraybackslash}p{0.27\linewidth} >{\raggedright\arraybackslash}X@{}}
\toprule
\tabhead{Mismatch} & \tabhead{Source} & \tabhead{Correction} \\
\midrule
\grouprow{3}{Infrastructure layer} \\
Weight version & asynchronous rollout & synchronization, bounded staleness \citep{fu2025areal,piche2025pipelinerl} \\
Engine numerics & low precision, kernels & batch-invariant kernels \citep{he2025nondeterminism}, FP16 training \citep{qi2025fp16} \\
Sampling and reporting & top-p, processed log-probabilities & report the drawn distribution \\
Sequence identity & retokenization & pass token ids to the trainer \\
Response selection & filtering, partial rollouts & outcome-independent batching \\
\midrule
\grouprow{3}{Algorithm layer} \\
Policy lag $\ell_t$ & weight version, grows with the update & decoupled PPO \citep{hilton2022batch}, M2PO \citep{Zheng2025ProsperityBC}, VCPO \citep{Huang2026StableAV} \\
Numerical discrepancy $e_t$ & engine numerics, approx.\ fixed in training & truncated importance sampling \citep{yao2025offpolicy}, score centering \citep{marek2026score} \\
\bottomrule
\end{tabularx}
\end{table}

\citet{yao2025offpolicy} reported that vLLM and FSDP give different token probabilities at identical weights and introduced truncated importance sampling, and \citet{liu2025speed} traced the gap to engine numerics, quantization and kernels and proposed sequence-level variants. \citet{zheng2025stabilizing} showed that the token-level objective approximates the sequence-level one only when both the numerical gap and the staleness are small. Engineering work reduces the numerical discrepancy through batch-invariant kernels \citep{he2025nondeterminism}, deterministic inference across tensor-parallel sizes \citep{Zhang2025DeterministicIA}, FP16 training \citep{qi2025fp16}, an FP32 language-model head \citep{minimax2025m1} and router replay for mixture-of-experts models \citep{ma2025r3}. \citet{zhong2026diagnosing} isolate the numerical discrepancy against a bitwise-aligned reference engine, and \citet{yu2026rethinking} model it as a displacement of the logits and calibrate the importance weight accordingly. For quantized samplers in synchronous training, AIS and QaRL weight the discrepancy factor separately from the PPO ratio \citep{Zhou2026AISAI,Gu2026QaRLRQ}, and \citet{Chen2026TowardsFP} calibrate the clip range to the quantization noise. QeRL instead trains with a quantized sampler and uses its quantization noise for exploration \citep{Huang2025QeRLBE}. QUADS aligns the quantization errors of the sampler and the trainer for mixture-of-experts models in 4-bit floating point \citep{Zhuge2026QUADSSN}.

\citet{marek2026score} trace the instability under mismatch to the drift, the term of the policy gradient that \cref{sec:setup} defines, and remove it by subtracting the sampler's mean score at each prefix. In their experiments the outcome of each correction depends on the mismatch source. With a quantized sampler, score centering alone trains best, and with a sampler synchronized every 64 steps, importance sampling with centering trains while score centering alone falls behind. PPO survives staleness and collapses under quantization and under synthetic weight noise. The authors attribute the stale case to the covariance that score centering takes under the sampler and the survival of PPO to a clip designed for ratios from policy movement. Our account places the sampler inside a training loop, since the sampler is a biased copy of the trainer that synchronizes back to it. \Cref{sec:sources} separates the part of the mismatch that moves with the trainer from the part that stays fixed, and \cref{sec:loop} follows the remainder after centering for the moving part through training.

\subsection{Off-policy correction and staleness}
Truncated and clipped importance weights have a long history in off-policy reinforcement learning \citep{ionides2008truncated,munos2016retrace,espeholt2018impala}, and ACER combines truncation with a bias-correction term over the full action set \citep{wang2017acer}. Asynchronous systems bound the staleness and weight each token by its ratio to the logged rollout policy \citep{fu2025areal,piche2025pipelinerl,noukhovitch2025asynchronous}, decoupled PPO separates a proximal policy from the behavior policy \citep{hilton2022batch}, and GSPO clips the ratio at the sequence level \citep{zheng2025gspo}. \citet{guan2026missing} show that the importance ratio factors into a training-inference discrepancy and a policy staleness and that this factorization needs the trainer's log-probabilities at the old weights. \citet{song2026scaling} trace the collapse of asynchronous GRPO to a staleness-induced update bias that positive feedback across synchronization periods sustains, and COPC couples the correction of the policy ratio with that of the advantage estimate \citep{hu2026copc}. Adaptive schemes control the update through statistics of the ratio. For example, M2PO bounds the second moment of the importance weights \citep{Zheng2025ProsperityBC}, VCPO scales the learning rate by the effective sample size \citep{Huang2026StableAV}, and \citet{guan2026missing} reset the proximal reference when the share of unmasked tokens falls below 0.9. Other studies act on the update itself in three ways \citep{chang2026muoneq}. GAC finds that successive gradients of asynchronous training stay aligned and projects the update off the stale-aligned direction \citep{Xu2026GACSA}, SAT shrinks one side of the PPO trust region with the log-ratio as a proxy for staleness \citep{Yang2026StaleBS}, and GMC-GRPO caps the importance mass of each group and bounds the bias of trajectory-level weighted estimators \citep{He2026AsynchronousLP}. GAC, SAT and GMC-GRPO analyze the update without centering and do not separate the policy lag from the numerical discrepancy. The discrepancy-lag correction of \cref{sec:method} builds on the factorization of \citet{guan2026missing}, gives each factor its own correction according to whether it enters the training loop and uses the second moment of the lag ratio to resynchronize the sampler's weights.

\section{Setup}\label{sec:setup}

\subsection{Sampler, trainer and corrections}
The trainer at weights $\theta$ has next-token distribution $p_\theta$, and the sampler draws responses $y$ from $q$. Each sampled token $y_t$ at prefix $h_t$ has score $s_t = \nabla_\theta \log p_\theta(y_t\mid h_t)$ and advantage $A_t$, the reward minus the group mean \citep{shao2024deepseekmath,ahmadian2024back,li2026right}. At a fixed prefix we write $p_v$ and $q_v$ for the probabilities of token $v$, $s$ for the score and $a$ for the expected advantage after the token \citep{li2026oppo}, both as functions of $v$. On policy the mean score is zero because the probabilities sum to one, so an advantage that is constant at the prefix has zero expected contribution. Under the sampler the update at the prefix splits into two terms \citep{marek2026score},
\begin{equation}
\begin{aligned}
\Ex_q[a\,s] &= \Ex_q[a]\,\Ex_q[s] + \mathrm{Cov}_q(a, s), \\
\Ex_q[s] &= \Ex_q[s] - \Ex_p[s] = -\nabla_\theta\,\mathrm{KL}(q\,\|\,p_\theta).
\end{aligned}
\label{eq:drift}
\end{equation}
We write $\bar a = \Ex_q[a]$ for the expected advantage of the prefix and $\bar s = \Ex_q[s]$ for the sampler's mean score. The first term $\bar a\,\bar s$ is the drift, a term that depends on the advantages only through their mean. By the second line the drift is a gradient step on the KL from the sampler to the trainer with step size $\bar a$. As a consequence, the drift distills the trainer toward the sampler on prefixes with positive expected advantage and pushes it away on prefixes with negative expected advantage. Group centering sets the mean advantage of each prompt to zero, whereas prefixes on the way to a correct answer have positive expected advantage and prefixes after an error have negative expected advantage. The drift is therefore nonzero on the prefixes that carry the learning signal.

The corrections differ in how they treat the two terms. The policy gradient averages $A_t s_t$ over the sampled tokens. Importance sampling multiplies each term by the truncated ratio $\min(p/q, 2)$. The untruncated ratio would cancel the drift, but its truncation on rare tokens brings part of the drift back. Score centering replaces $s_t$ by $s_t - \bar s_t$, so that the centered update equals the covariance term of \cref{eq:drift} taken under $q$. The sampler reports its log-probabilities on its top-128 tokens $H$. As in verl \citep{sheng2024hybridflow}, the tail takes the trainer's probabilities scaled by the ratio of the two tail masses, $\kappa = (1 - \sum_H q_v)/(1 - \sum_H p_v)$. Since the full-vocabulary mean score under $p$ is zero, the mean score that score centering subtracts reduces to $\hat s_t = \sum_{v\in H}(q_v - \kappa\, p_v)\,s_v$. Importance sampling with centering weights each token as importance sampling does and subtracts the sampler's mean of the weighted score. PPO clips the ratio of $p_\theta$ to a reference policy, which is either the trainer at the sampler's weights or the trainer before each batch \citep{hilton2022batch,fu2025areal}.

\subsection{Mismatch sources and training}
The reference sampler runs vLLM in BF16 with CUDA graphs and yields on-policy training. The quantized sampler stores the weights in FP8 \citep{micikevicius2022fp8} with $128\times128$ blocks, quantizes the activations dynamically and keeps an FP8 KV cache. The stale sampler runs in BF16 and synchronizes every 64 steps, so within the 40 training steps it keeps the initial weights $\theta_0$ and the accumulated update is $\Delta = \theta - \theta_0$. The combined source quantizes the stale weights $\theta_0$ in the same way. The four samplers draw at temperature one without truncation and pass token ids to the trainer, so the sampling, reporting and sequence mismatches of \cref{tab:directions} vanish.

We train Qwen3-0.6B and Qwen3-1.7B \citep{yang2025qwen3} in non-thinking mode on Countdown with rewards $\pm1$, 64 prompts with eight samples per step and SGD \citep{chang2025convergence} at learning rate $10^{-2}$, and train each configuration three times for 40 steps. These short trainings, together with replay of the sampler on shared checkpoints, keep the measurements on trajectories that have not collapsed. The deviation of a trained model is its full-vocabulary KL to the on-policy model trained with the same configuration on fixed prefixes from $\theta_0$, averaged over the three pairs and divided by the noise floor, the mean KL between two independently trained on-policy models. Test accuracy is measured on 64 Countdown problems outside the training set with eight samples each. \Cref{app:notation} collects the notation.

\section{Numerical Discrepancy and Self-Induced Policy Lag}\label{sec:sources}

\subsection{Source dependence of the corrections}
\begin{figure}[t]
\centering
\includegraphics[width=\linewidth]{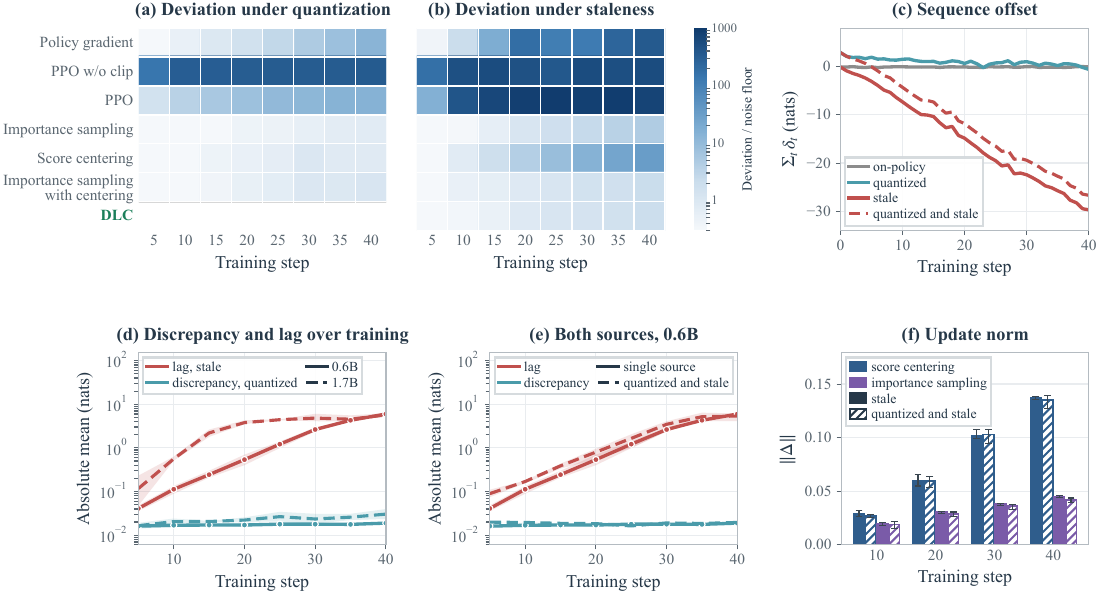}
\caption{\textbf{Deviation of the corrections and the growth of the discrepancy and the lag.} The colour of each cell gives the deviation of a Qwen3-0.6B run as a multiple of the noise floor on a log scale, averaged over the three runs, and \cref{tab:source} lists the values at step 40. Without a policy lag DLC reduces to score centering, so DLC has a run only under the two stale sources. Under quantization the corrections stay near the floor (a), whereas under staleness score centering leaves it and importance sampling with centering and DLC stay closest to it (b). The sequence offset stays constant under quantization and grows under staleness (c). The numerical discrepancy of the quantized sampler stays near 0.02 nats per token while the policy lag of a stale sampler grows by two orders of magnitude on both model sizes (d). With both sources combined, the numerical discrepancy keeps the size it has under quantization alone, and the lag reaches the stale value by step 40 (e). The accumulated update of the combined source follows that of the stale runs, with the hatched bars for the combined source and the vertical lines spanning the three runs (f).}
\label{fig:source}
\end{figure}

\Cref{tab:source} shows that our 40-step trainings reproduce the source dependence of the corrections. Under quantization, importance sampling and score centering, alone or combined, stay at the noise floor while the policy gradient reaches 13 times the floor, and score centering matches the 46.5\% test accuracy of on-policy training. Under staleness, score centering reaches 32 times the floor and ends at 12.2\%, while importance sampling with centering stays at 2.1 times the floor and ends at 32.0\%. Our PPO runs take the trainer before each batch as the reference of the clip and reach 699 times the floor under staleness. In the experiments of \citet{marek2026score}, PPO trains under staleness, and \cref{sec:range} traces this difference to the reference policy. Panels a and b of \cref{fig:source} follow these deviations over training.

\begin{table}[!htb]
\centering
\caption{\textbf{The corrections under each mismatch source on Qwen3-0.6B.} The deviation is the KL to on-policy training as a multiple of the noise floor at steps 20 and 40, and lower values are better. Test accuracy is in percent at step 40, the last training step, and higher values are better. PPO uses the trainer before each batch as its reference, and DLC is the correction of \cref{sec:method}. Among the baselines and DLC, the best value in each column is in bold and the second best is underlined, and a dash marks a correction without a run under that source.}
\label{tab:source}
\small
\setlength{\tabcolsep}{4pt}
\begin{tabular}{@{}l *{3}{ccc}@{}}
\toprule
 & \multicolumn{3}{c}{\tabhead{Quantized sampler}} & \multicolumn{3}{c}{\tabhead{Stale sampler}} & \multicolumn{3}{c}{\tabhead{Quantized and stale}} \\
\cmidrule(lr){2-4}\cmidrule(lr){5-7}\cmidrule(l){8-10}
 & \multicolumn{2}{c}{\tabhead{\begin{tabular}[b]{@{}c@{}}Deviation\\at step\end{tabular}}} &  & \multicolumn{2}{c}{\tabhead{\begin{tabular}[b]{@{}c@{}}Deviation\\at step\end{tabular}}} &  & \multicolumn{2}{c}{\tabhead{\begin{tabular}[b]{@{}c@{}}Deviation\\at step\end{tabular}}} &  \\
\cmidrule(lr){2-3}\cmidrule(lr){5-6}\cmidrule(lr){8-9}
\tabhead{Correction} & \tabhead{20} & \tabhead{40} & \tabhead{Accuracy} & \tabhead{20} & \tabhead{40} & \tabhead{Accuracy} & \tabhead{20} & \tabhead{40} & \tabhead{Accuracy} \\
\midrule
\grouprow{10}{Reference} \\
\textcolor{gray}{Initial model $\theta_0$} & \textcolor{gray}{--} & \textcolor{gray}{--} & \textcolor{gray}{8.2} & \textcolor{gray}{--} & \textcolor{gray}{--} & \textcolor{gray}{8.2} & \textcolor{gray}{--} & \textcolor{gray}{--} & \textcolor{gray}{8.2} \\
\textcolor{gray}{On-policy training} & \textcolor{gray}{1.00} & \textcolor{gray}{1.00} & \textcolor{gray}{46.5} & \textcolor{gray}{1.00} & \textcolor{gray}{1.00} & \textcolor{gray}{46.5} & \textcolor{gray}{1.00} & \textcolor{gray}{1.00} & \textcolor{gray}{46.5} \\
\midrule
\grouprow{10}{Baselines} \\
Policy gradient & 1.59 & 12.7 & 26.4 & 178 & 345 & 0.2 & 45.0 & 107 & 0.2 \\
PPO w/o clip & 314 & 336 & 0.0 & 466 & 534 & 0.0 & -- & -- & -- \\
PPO & 7.52 & 14.3 & 26.2 & 817 & 699 & 0.0 & -- & -- & -- \\
Importance sampling & \textbf{0.43} & \textbf{0.74} & 43.2 & \underline{1.16} & 4.50 & 20.9 & 1.25 & 3.91 & 27.1 \\
Score centering & \underline{0.53} & \underline{0.80} & \textbf{46.5} & 3.91 & 31.7 & 12.2 & 4.19 & 28.1 & 18.4 \\
Importance sampling with centering & \underline{0.53} & 1.11 & \underline{46.0} & \textbf{0.86} & \underline{2.10} & \textbf{32.0} & \underline{1.20} & \underline{2.61} & \textbf{30.5} \\
\midrule
\rowcolor{oursbg}
\textbf{DLC (ours)} & -- & -- & -- & \textbf{0.86} & \textbf{1.93} & \underline{31.7} & \textbf{0.96} & \textbf{2.17} & \underline{30.3} \\
\bottomrule
\end{tabular}
\end{table}

\subsection{Exact split of the sampled-token error}
The untruncated ratio and centering both remove the drift at a prefix, so the corrections differ in what they leave, and the mismatch enters this remainder through the sampled-token error \mbox{$\delta_t = \log p_\theta(y_t\mid h_t) - \log q(y_t\mid h_t)$}. The sampler differs from the trainer in its weights and in its engine, and a third log-probability at the sampler's weights $\old$ with the trainer's numerics changes one attribute at a time and thereby splits the error exactly,
\begin{equation}
\delta_t = \underbrace{\log p_\theta(y_t\mid h_t) - \log p_{\old}(y_t\mid h_t)}_{\ell_t} + \underbrace{\log p_{\old}(y_t\mid h_t) - \log q(y_t\mid h_t)}_{e_t}.
\label{eq:split}
\end{equation}
We call $\ell_t$ the policy lag and $e_t$ the numerical discrepancy. The split is the logarithm of the factorization of the importance ratio into a policy-staleness factor and a training-inference discrepancy factor at the sampler's weights \citep{hilton2022batch,fu2025areal,guan2026missing}, and the trainer's reference forward pass at $\old$ supplies its middle term.

The two terms follow different dynamics during training on both model sizes. The error of the quantized sampler is numerical discrepancy, about 0.02 nats per token throughout training, and the stale error is policy lag, with numerical discrepancy below $10^{-3}$, and the quantization error and the lag separate by step 5 in panel d of \cref{fig:source}. Summed over a response, the quantization error gives a constant sequence offset while the magnitude of the stale offset grows by about 0.7 nats per step, as panel c shows. To first order the lag of a token is the accumulated update projected on its score, $\ell_t \approx \Delta^\top s_t$, so the policy lag is self-induced and grows only because the trainer moves. Its mean $\mu$ under the sampler is minus the per-token KL from the sampler's weights to the trainer when the numerical discrepancy is zero. The first-order term has mean zero, so $\mu$ measures the spread of the per-token lag at second order.

The numerical discrepancy does not depend on $\Delta$, so it biases each step by the same amount without feeding into later errors, and centering removes its first-order bias, as the quantization columns of \cref{tab:source} show. The runs that combine quantization with staleness test this split. Their numerical discrepancy keeps the value it has under quantization alone in panel e of \cref{fig:source}, and their accumulated update stays within ten percent of that of the stale runs for both score centering and importance sampling in panel f. Their lag starts up to twice as large and reaches the stale value by step 40, and their deviations keep the stale order.

\section{Training Loop of the Policy Lag}\label{sec:loop}

\subsection{Links of the loop}
\begin{figure}[!htb]
\centering
\includegraphics[width=\linewidth]{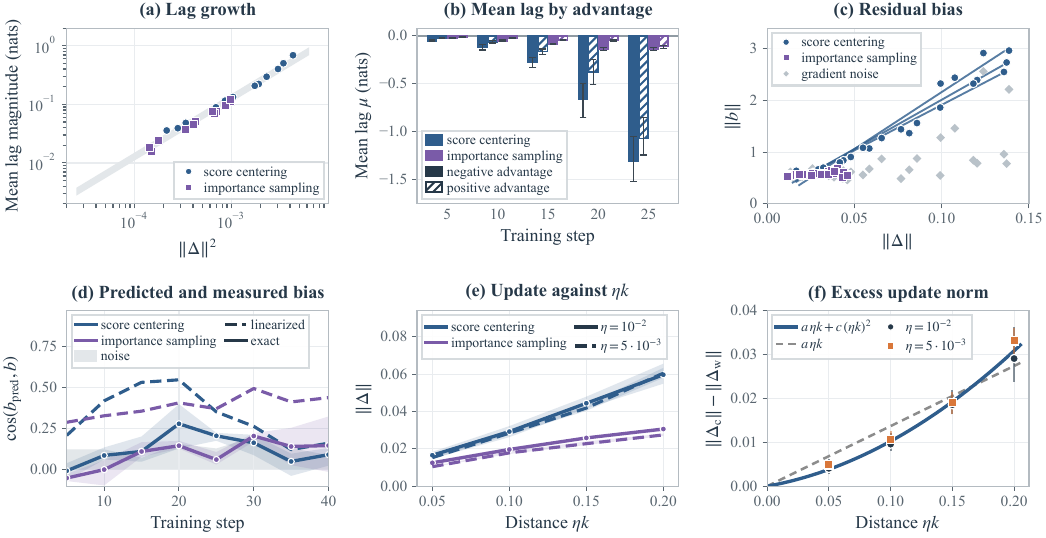}
\caption{\textbf{Measurements of the three links of the loop.} We measure the three links under a stale sampler on Qwen3-0.6B. The mean lag grows with $\|\Delta\|^2$ at the same rate under score centering and importance sampling, inside the grey band of 110 to 150 times $\|\Delta\|^2$ (a), and tokens with negative advantage carry the larger lag only under score centering (b). The residual bias of score centering grows linearly in $\|\Delta\|$ while that of importance sampling stays at the gradient noise (c). The predicted bias aligns with the measured one above the gradient noise (d), the update depends on the learning rate only through $\eta k$ (e), and the excess update norm of score centering over importance sampling follows the quadratic loop fit (f).}
\label{fig:loop}
\end{figure}

Under staleness the infrastructure keeps the sampler at $\theta_0$, so the version gap equals the accumulated update, and a training step $k$ runs through three links. In lag growth the accumulated update $\Delta_k$ sets the lag of each sampled token, in the residual bias the lag sets the bias $b_k$ that the correction leaves, and in bias accumulation the bias enters the next update,
\begin{equation}
\Delta_{k+1} = \Delta_k + \eta\,(g_k + b_k) \approx (I - \eta M)\,\Delta_k + \eta\,g_k,
\label{eq:loop}
\end{equation}
with learning rate $\eta$ and on-policy gradient $g_k$. The second step uses the fact that the bias is linear in the accumulated update to first order, $b_k \approx -M\Delta_k$, where $M$ collects the per-token weights that the correction leaves and \cref{eq:bias} gives its form. Along an eigenvector of $M$ with eigenvalue $\lambda$, a step multiplies the bias-driven part of $\Delta$ by $1 - \eta\lambda$, so that part grows where $\lambda < 0$ until the next synchronization resets $\Delta$ to zero. Up to step 20 the magnitude of the mean lag stays between 110 and 150 times $\|\Delta\|^2$ under both corrections in panel a of \cref{fig:loop}, so score centering and importance sampling produce the same lag per unit of movement and differ in how far $\Delta$ moves.

\subsection{Residual bias of each correction}
By \cref{eq:drift} centering removes the drift exactly and leaves $\mathrm{Cov}_q(a, s)$, while the on-policy update at the same prefix is $\mathrm{Cov}_p(a, s)$ because the on-policy mean score is zero. Over a response the remaining bias follows from replacing one distribution at a time. Let $\mu^\pi_t$ be the distribution of the prefix $h_t$ under $\pi$, $a^\pi_t$ the expected advantage after $y_t$ when $\pi$ generates the rest of the response and $\hat s_t$ the head estimate of $\bar s_t$ from \cref{sec:setup}. Score centering subtracts $\hat s_t$, so at each prefix its update equals $\mathrm{Cov}_q(a^q_t, s_t) + \Ex_q[a^q_t]\,(\bar s_t - \hat s_t)$, and its bias $b_\mathrm{c}$ is exactly
\begin{equation}
\begin{aligned}
b_\mathrm{c} &= \sum_t \Big[\Ex_{\mu^q_t}\Big(\sum_v q_v\,a^q_t(v)\,\big(s_v - \hat s_t\big)\Big) - \Ex_{\mu^p_t}\Big(\sum_v p_v\,a^p_t(v)\,s_v\Big)\Big] \\
&= \sum_t \Big[\Ex_{\mu^q_t}\mathrm{Cov}_q(a^q_t, s_t) - \Ex_{\mu^p_t}\mathrm{Cov}_p(a^p_t, s_t) + \Ex_{\mu^q_t}\Ex_q[a^q_t]\,(\bar s_t - \hat s_t)\Big] \\
&= \sum_t \Big[\big(\Ex_{\mu^q_t} - \Ex_{\mu^p_t}\big)\mathrm{Cov}_q(a^q_t, s_t) + \Ex_{\mu^p_t}\mathrm{Cov}_q(a^q_t - a^p_t, s_t) \\
&\qquad\quad + \Ex_{\mu^p_t}\big(\mathrm{Cov}_q(a^p_t, s_t) - \mathrm{Cov}_p(a^p_t, s_t)\big) + \Ex_{\mu^q_t}\Ex_q[a^q_t]\,(\bar s_t - \hat s_t) \Big],
\end{aligned}
\label{eq:exact}
\end{equation}
where the last step adds and subtracts $\Ex_{\mu^p_t}\mathrm{Cov}_q(a^q_t, s_t)$ and $\Ex_{\mu^p_t}\mathrm{Cov}_q(a^p_t, s_t)$ and uses the linearity of the covariance in its first argument.
In the order of the last two lines, we call the four terms the prefix term, the continuation term, the token term and the tail term. The tail term depends on the head $H$ that the sampler reports, so the infrastructure sets it. The other three terms come from the sampled-token error $\delta_t = \ell_t + e_t$. The numerical discrepancy gives them the same size at each step, and the policy lag gives them a part that grows with $\Delta$.

Under quantization the sequence offset stays at 8.9 nats on long responses while the residual bias stays at the gradient noise, so the size of the offset alone does not bias the update. Under staleness, score centering leaves the floor near step 12, where the growing offset reaches the same size, so the deviation follows the covariance of the offset with the gradient that the policy lag supplies.

The importance weight of the prefix $h_t$ from the sampler to the trainer is $e^{S_t}$ with $S_t = \sum_{t'<t}\delta_{t'}$, and the continuation term has the same form with the error summed after $t$. For the token term, a first-order expansion of the sampler in the lag shows that the centered update at a prefix differs from the on-policy update by the lag weighted with the deviation of the advantage from its on-policy mean $\bar A_t = \Ex_p[a]$ at the prefix, as \cref{app:derivation} derives step by step. The ratio of importance sampling cancels the token term on tokens whose ratio stays below the truncation at 2, and the truncated tokens keep the excess of their ratio over 2 as a weight. Adding the linearized prefix term and then substituting the first-order lag $\Delta^\top s_t$ of each token under staleness gives
\begin{equation}
\begin{aligned}
b_\mathrm{c} &\approx -\sum_t \Ex\Big[\big((A_t - \bar A_t)\,\ell_t + A_t S_t\big)\, s_t\Big] \approx -M_\mathrm{c}\,\Delta, \qquad
M_\mathrm{c} = \sum_t \Ex\Big[s_t\Big((A_t - \bar A_t)\,s_t + A_t\sum_{t'<t} s_{t'}\Big)^{\!\top}\Big], \\
b_\mathrm{w} &\approx -\sum_t \Ex\Big[A_t\big(S_t + \max(e^{\ell_t} - 2,\, 0)\big)\, s_t\Big] \approx -M_\mathrm{w}\,\Delta, \qquad
M_\mathrm{w} = \sum_t \Ex\Big[A_t\, s_t \sum_{t'<t} s_{t'}^\top\Big],
\end{aligned}
\label{eq:bias}
\end{equation}
where $b_\mathrm{w}$ is the bias of importance sampling and its second step takes the tokens inside the validity range, on which the truncation weight is zero. Both biases are policy gradients with per-token weights built from the lag, and these weights give $M$ of \cref{eq:loop}. Score centering keeps the token term $(A_t - \bar A_t)\ell_t$, which importance sampling removes. The prediction below keeps the token and prefix terms.

The measured residual bias agrees with these forms for both corrections. We measure $b$ as the corrected gradient minus the on-policy gradient on the same batch and compare it with the gradient noise, the difference of two independent on-policy gradients. Under score centering, $\|b\|$ grows linearly in $\|\Delta\|$ with $R^2$ of 0.95 to 0.99 per run and reaches 2.5 times the gradient noise by step 40, whereas under importance sampling $\|b\|$ stays within the noise, as panel c of \cref{fig:loop} shows. The weights of \cref{eq:bias} also predict $b$ from one backward pass on the same batch. Up to step 20 the prediction aligns with the measured bias at a cosine of up to 0.55, against at most 0.07 for the gradient noise, mainly through the prefix term in panel d. The linearized prefix weight $S_t$ grows with the position while the exact weight $e^{S_t} - 1$ stays above $-1$, so the prediction overstates the norm. With exact weights the overstatement falls below eight times and the cosine stays above the noise, as \cref{app:exact} shows.

\subsection{Loop gain}
Through bias accumulation the bias changes the lag of token $t$ by $\eta\,s_t^\top b_k$ beyond the gradient. Projected through $\ell_t \approx \Delta^\top s_t$, the token term of score centering changes each lag per step by \mbox{$-\eta\,\Ex_{t'}\big[(s_t^\top s_{t'})(A_{t'} - \bar A_{t'})\,\ell_{t'}\big]$} over the tokens $t'$ of the batch. Each token with an advantage below its prefix mean therefore enlarges the lag magnitude on the tokens whose scores align with its own, itself included, and this enlargement produces the negative eigenvalues of $M$. Grouping the sampled tokens by advantage confirms this sign, since under score centering the mean lag at step 20 is $-0.66$ on tokens with negative advantage and $-0.37$ on tokens with positive advantage, while under importance sampling the two groups agree in panel b of \cref{fig:loop}.

The learning rate separates the loop from a bias that is fixed in $\Delta$. Unrolling \cref{eq:loop} to second order, as \cref{app:loop} details, gives $\Delta_k \approx \eta k\,\bar g - \tfrac12(\eta k)^2 M\bar g$ for a mean on-policy gradient $\bar g$, which depends on $\eta$ and $k$ only through the distance $\eta k$. Under a bias fixed in $\Delta$ the excess update norm $E = \|\Delta_\mathrm{c}\| - \|\Delta_\mathrm{w}\|$ of score centering over importance sampling is linear in $\eta k$, and the loop adds a quadratic term. In panel e of \cref{fig:loop} the runs at half the learning rate fall on the curves of the original ones, and in panel f the fit $E = a\,\eta k + c\,(\eta k)^2$ over both learning rates gives $c = 0.52$ at $R^2 = 0.987$, against 0.90 for the linear term alone. The quadratic coefficient is positive in each of the six fits by learning rate and run. The dependence on $\eta k$ also agrees with the stability boundary of asynchronous GRPO, which lies at a constant product of synchronization interval and learning rate \citep{song2026scaling}. Beyond step 20 the lag under score centering passes one nat, and the first-order loop ends at this size. The lag then spreads over the response, so that the 1\% of tokens with the largest lag carry 78\% of the summed lag at step 5 and 11\% at step 40.

\section{Validity Range of Importance Weighting}\label{sec:range}

\subsection{Tail of the ratio}
\begin{figure}[!htb]
\centering
\includegraphics[width=\linewidth]{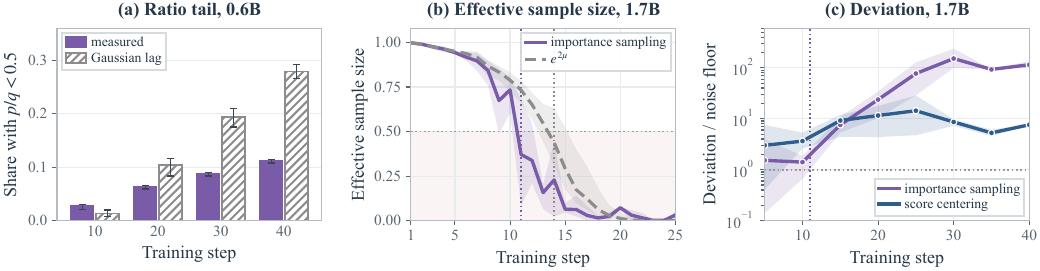}
\caption{\textbf{Validity range of importance weighting under staleness.} Fewer tokens fall below a ratio of 0.5 than a Gaussian lag predicts (a). On Qwen3-1.7B the effective sample size of importance sampling therefore drops below one half at step 11, three steps before the Gaussian value, with the purple and grey dotted lines at the two crossings (b), and the deviation of importance sampling leaves the floor from step 11 on (c).}
\label{fig:range}
\end{figure}

Ratio weighting cuts the loop with a weight that itself depends on the lag. The importance sampling ratio of a token under staleness is $\rho_t = e^{\ell_t}$, with mean one under $q$, so a negative mean lag comes with a spread of ratios and an effective sample size $1/\Ex_q[\rho_t^2]$ as a fraction of the batch, which equals $e^{2\mu}$ for a Gaussian lag by the derivation of \cref{app:ess}. On Qwen3-0.6B the mean lag under importance sampling stays below 0.3 nats in magnitude through step 40. At step 40 a Gaussian lag with the measured $\mu$ predicts that 28\% of the tokens have a ratio below 0.5, whereas 11\% do in panel a of \cref{fig:range}, so most tokens stay near ratio one and the lag concentrates on few of them \citep{vehtari2024pareto}. On Qwen3-1.7B the updates per step are larger, and the training logs give the second moment of the ratio at each step. The measured effective sample size agrees with the Gaussian value at step 10 and falls below one half at step 11, at a mean lag of $-0.15$ nats per token. The Gaussian value crosses one half three steps later, at step 14 in panel b.

The validity range of ratio weighting therefore ends earlier than $e^{2\mu}$ indicates, at a mean lag of $-0.15$ nats per token on Qwen3-1.7B against the Gaussian $-0.35$. Past this range the truncated tokens bias the update, and on Qwen3-1.7B the deviation of importance sampling leaves the floor at the step where the effective sample size crosses one half and grows to 152 times the floor by step 30, as panel c of \cref{fig:range} shows. Under a constant policy lag of two to twelve steps, \citet{Huang2026StableAV} observe the same order for sequence-level importance sampling, whose effective sample size collapses before the KL to the sampler grows.

\subsection{Reference policy of PPO}
The same derivation fixes the role of the PPO reference policy. With the reference at the sampler's weights, the clipped ratio is $e^{\ell_t}$, so the clip bounds the token term as the importance sampling ratio does and leaves the numerical discrepancy outside the ratio as drift. With the reference at the trainer before each batch, the policy lag stays outside the ratio, and the clip bounds only the movement between the updates within a batch. Our PPO runs take four updates per batch, and PPO without clip takes the same updates with the clip removed. Under staleness, PPO with the batch reference deviates at least as far as PPO without clip, at 699 against 534 times the floor in \cref{tab:source}, while the sampler-weight reference of DLC stays at 1.9. Under quantization, PPO stays at the level of the policy gradient with one update per batch, so the clip removes the cost of the extra updates and leaves the drift of the numerical discrepancy. \citet{marek2026score} attribute this asymmetry to a clip designed for ratios that come from policy movement, and \citet{Chen2026TowardsFP} observe that quantization noise in the ratio pushes negative-advantage tokens outside the clip range. The position of the reference policy therefore accounts for the asymmetry. With the sampler's weights as reference, the policy lag lies inside the ratio of PPO and the numerical discrepancy outside it, so the clip contains the lag under staleness and leaves the drift of the discrepancy under quantization. Our runs test the batch reference directly and the sampler reference through DLC, which shares its clipped ratio.

\section{Discrepancy-Lag Correction}\label{sec:method}

\subsection{Method}
The discrepancy-lag correction (DLC) assigns each term of \cref{eq:exact} to the layer that can remove it. Inside the algorithm, the policy lag needs its token term removed and the numerical discrepancy needs its drift removed. The lag that the algorithm cannot reach, in the prefix and continuation terms and on the tokens past the validity range, goes back to the infrastructure through a resynchronization of the sampler. The trainer runs one extra forward pass at the sampler's weights $\old$ with its own numerics. On each sampled token this pass gives $\ell_t$ and $e_t$ through \cref{eq:split}, and on the sampler's top-128 tokens $H$ it gives the log-probabilities of $p_{\old}$ that the centering term below needs.

The policy lag lies inside the loop, so its correction has to remove the token term $(A_t - \bar A_t)\ell_t$ of \cref{eq:bias}, which sets the loop gain. Ratio weighting removes this term, and for that purpose the lag factor of the ratio suffices. DLC therefore clips the lag ratio $\rho_t = e^{\ell_t}$ with the PPO clip against the sampler's weights, so that token $t$ enters the update with the weight $m_t\rho_t$. The clip indicator $m_t$ drops a token with positive advantage when its ratio exceeds $1 + \varepsilon$ and a token with negative advantage when its ratio leaves the interval from $1 - \varepsilon$ to $c$, with $\varepsilon = 0.2$ and the dual-clip bound $c = 3$ of verl. Inside the clip range this weight is the lag ratio, as for importance sampling under staleness, so it removes the token term to the same order, and \cref{app:slc} shows that the lag enters the token term of DLC only through its product with the numerical discrepancy.

The sampled tokens come from $q$, so the numerical discrepancy between $q$ and $p_{\old}$ remains in the clipped update. At a prefix with a constant advantage $A$ the clipped update has the expectation $A\sum_v q_v m_v \rho_v s_v$ under the sampler, which is nonzero for $q \neq p_{\old}$ and is the drift of \cref{eq:drift} for this estimator. Weighting by $p_{\old}/q$ would cancel this drift only without truncation, so DLC instead subtracts the expectation itself at each sampled token, as score centering subtracts $\bar s$, and the subtracted term then follows the clip. The sampler reports $q$ only on its head $H$, and on the tail DLC takes $q_v = \kappa\,p_{\old}(v)$ with the ratio of the tail masses $\kappa = (1 - \sum_H q_v)/(1 - \sum_H p_{\old}(v))$ and leaves the clip inactive, so that $q_v m_v \rho_v = \kappa\,p_\theta(v)$ there. The update of DLC then follows in three steps,
\begin{equation}
\begin{aligned}
\hat g_\mathrm{DLC} &= \sum_t A_t\Big(m_t\,\rho_t\,s_t - \sum_{v} q_v\,m_v\,\rho_v\,s_v\Big) \\
&= \sum_t A_t\Big(m_t\,\rho_t\,s_t - \sum_{v\in H} q_v\,m_v\,\rho_v\,s_v - \kappa\sum_{v\notin H} p_\theta(v)\,s_v\Big) \\
&= \sum_t A_t\,\big(m_t\,\rho_t\,s_t - C_t\big), \qquad C_t = \sum_{v\in H}\big(q_v\,m_v\,\rho_v - \kappa\,p_\theta(v)\big)\,s_v .
\end{aligned}
\label{eq:slc}
\end{equation}
The first line subtracts the sampler's expectation of the clipped score at prefix $h_t$, the second line splits this expectation into the head that the sampler reports and the estimated tail, and the third line uses $\sum_v p_\theta(v)\,s_v = 0$ to write the tail sum as minus the same sum over the head. The two terms of the last line divide the work between the numerical discrepancy and the policy lag. The weighted score $m_t\rho_t s_t$ carries the lag ratio and removes the token term of the policy lag inside the clip range. The centering term $C_t$ is the sampler's mean of this weighted score, so under the sampler a constant advantage at a prefix contributes zero up to the tail estimate, as it does on policy, and the drift of the numerical discrepancy vanishes. Only the head log-probabilities of the sampler and of $p_{\old}$ enter $C_t$, and \cref{app:split} gives the loss that yields the update in one backward pass. \citet{guan2026missing} weight both factors of the ratio, and DLC weights the lag factor and centers the discrepancy factor.

The clip keeps the validity range of ratio weighting unchanged. Tokens whose lag ratio leaves the clip range drop out of the update, so their part of the token term returns, as it does for the truncated tokens of importance sampling. The share of these tokens grows as the lag ratio spreads, and \cref{sec:range} places the end of the range where the effective sample size falls to one half. DLC therefore copies the trainer's weights to the sampler when the effective sample size of the lag ratio falls below one half of the batch. The resynchronization removes the lag through the infrastructure, since it resets $\Delta$, the lag and the prefix offset to zero and moves the reference of the clip to the trainer's weights. Samplers that update their weights during generation keep this assignment, with the generating weights of each token as the reference of the clip and the admission of a response conditioned on its effective sample size in place of the resynchronization.

DLC reduces to the earlier corrections in two limits. With $\theta = \old$ the lag ratio is one and the clip is inactive, so $C_t$ becomes the mean score that score centering subtracts and DLC reduces to score centering. With $q = p_{\old}$ and an inactive clip, $\kappa = 1$ and $C_t$ vanishes, so DLC reduces to PPO with the sampler's weights as reference. Compared with importance sampling with centering, DLC adds one forward pass of the trainer at $\old$ per step and stores the same top-128 log-probabilities of the sampler. Its hyperparameters are the clip range, the head size and the resynchronization threshold, which the experiments keep at 0.2, 128 and one half. \Cref{tab:design} writes each correction as a per-token gradient, and \cref{alg:slc} lists one training step of DLC. The corrections differ mainly in the denominator of their ratio. The three importance sampling corrections and VCPO divide by $q$, so their weight and its truncation act on the numerical discrepancy as well as on the lag. PPO with sampler reference, M2PO and DLC divide by $p_{\old}$ and weight the lag alone, and of the three only DLC removes the drift of the numerical discrepancy by centering. PPO with batch reference divides by the trainer before the batch, so neither the discrepancy nor the lag enters its ratio. The adaptive controls differ in what they act on, since VCPO slows the update as the effective sample size falls, whereas importance sampling with resynchronization and DLC move the sampler to the trainer and reset the lag.

\begin{table}[!htb]
\centering
\caption{\textbf{Per-token gradients of the corrections.} The ratio denominator decides which component of the sampled-token error enters the weight, with $q$ for the discrepancy and the lag together and $p_{\old}$ for the lag alone. Here $w_t = \min(p_\theta/q,\, 2)$, $\rho_t = p_\theta/p_{\old}$, $r_t = p_\theta/p_{\theta_\text{batch}}$ for the trainer before each batch, and $m_t$ and $m^\text{b}_t$ are the clip indicators of \cref{eq:slc} on $\rho_t$ and $r_t$. The M2PO baseline masks tokens by $m^{\mathrm{M2}}_t$ until the second moment of $\rho_t$ is at most 0.04. VCPO weights response $i$ by $\bar w_i = \min\big(\prod_{t'} p_\theta/q,\, 8\big)$ and replaces its advantage by $\hat A_i$ with an off-policy baseline. In the last column the sampler resynchronizes when the effective sample size $1/\Ex_q[\rho_t^2]$ falls below one half, VCPO scales the learning rate by the inverse root of the second moment of $\bar w_i$, and $K$ counts the updates per batch.}
\label{tab:design}
\small
\renewcommand{\arraystretch}{1.25}
\setlength{\tabcolsep}{3.5pt}
\begin{tabular}{@{}>{\raggedright\arraybackslash}p{0.27\linewidth} l c c c@{}}
\toprule
\tabhead{Method} & \tabhead{Per-token gradient} & \tabhead{\begin{tabular}[b]{@{}c@{}}Ratio\\denominator\end{tabular}} & \tabhead{Centering} & \tabhead{\begin{tabular}[b]{@{}c@{}}Adaptive\\control\end{tabular}} \\
\midrule
\grouprow{5}{Corrections at each token} \\
Policy gradient & $A_t\,s_t$ & -- & \xmark & -- \\
Importance sampling & $w_t\,A_t\,s_t$ & $q$ & \xmark & -- \\
Score centering & $A_t\,(s_t - \bar s_t)$ & -- & \cmark & -- \\
Importance sampling with centering & $A_t\,\big(w_t\,s_t - \sum_v q_v w_v s_v\big)$ & $q$ & \cmark & -- \\
Importance sampling with resynchronization & $w_t\,A_t\,s_t$ & $q$ & \xmark & $1/\Ex_q[\rho_t^2] < \tfrac12$ \\
\midrule
\grouprow{5}{Reference policy of the PPO clip} \\
PPO with batch reference & $m^\text{b}_t\,r_t\,A_t\,s_t$ & $p_{\theta_\text{batch}}$ & \xmark & $K = 4$ \\
PPO with sampler reference & $m_t\,\rho_t\,A_t\,s_t$ & $p_{\old}$ & \xmark & -- \\
\midrule
\grouprow{5}{Methods for stale samplers} \\
M2PO & $m^{\mathrm{M2}}_t\,\rho_t\,A_t\,s_t$ & $p_{\old}$ & \xmark & -- \\
VCPO & $\bar w_i\,\hat A_i\,s_t$ & $q$ & \xmark & $\eta\,\big(\Ex[\bar w_i^2]\big)^{-1/2}$ \\
\midrule
\rowcolor{oursbg}
\textbf{DLC (ours)} & $A_t\,\big(m_t\,\rho_t\,s_t - C_t\big)$ & $p_{\old}$ & \cmark & $1/\Ex_q[\rho_t^2] < \tfrac12$ \\
\bottomrule
\end{tabular}
\end{table}

\begin{algorithm}[!htb]
\caption{Training step of DLC}
\label{alg:slc}
\small
\begin{algorithmic}[1]
\Require trainer weights $\theta$, sampler weights $\old$, advantages $A$ from the group rewards
\State Sample responses with the sampler at $\old$ and record $\log q$ on each sampled token and on the top-128 tokens $H$
\State Run the trainer's forward pass at $\old$ for $\log p_{\old}$ on the same tokens
\State Split each sampled-token error into the policy lag $\ell_t$ and the numerical discrepancy $e_t$ by \cref{eq:split}
\State Compute the effective sample size $1/\Ex_q[\rho_t^2]$ of the lag ratio $\rho_t = e^{\ell_t}$
\State Take one SGD step on the update $\sum_t A_t\,(m_t\,\rho_t\,s_t - C_t)$ with the centering term $C_t$ of \cref{eq:slc}
\If{the effective sample size is below one half of the batch}
\State Copy the trainer's weights to the sampler, so that $\old$ becomes $\theta$
\EndIf
\end{algorithmic}
\end{algorithm}

\subsection{Deviation and test accuracy}
\begin{figure}[!htb]
\centering
\includegraphics[width=\linewidth]{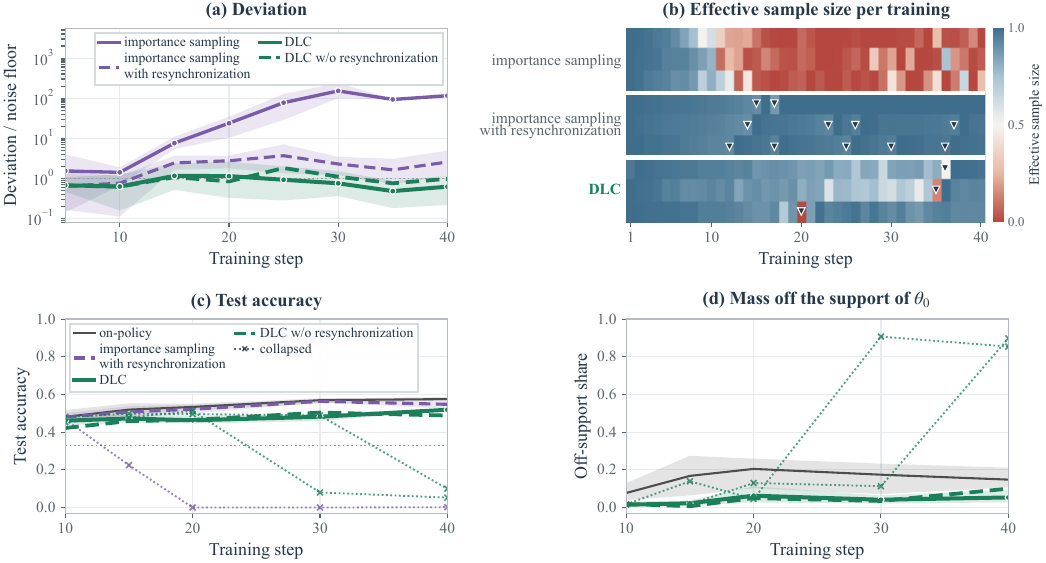}
\caption{\textbf{DLC against importance sampling, importance sampling with resynchronization and DLC without resynchronization on Qwen3-1.7B under staleness.} DLC stays at the floor and resynchronization brings importance sampling back near it (a), since each resynchronization, marked by a triangle in the row of its run, restores the effective sample size (b). The DLC models track on-policy training in test accuracy, whereas two runs of DLC without resynchronization and one of importance sampling with resynchronization fall below $\theta_0$ (c). The collapses without resynchronization coincide with a jump of the trainer's mass on tokens to which $\theta_0$ gives probability below $10^{-3}$, and this mass stays low under DLC (d).}
\label{fig:slc}
\end{figure}

On Qwen3-0.6B in \cref{tab:main} the DLC runs reach a deviation of 1.9 times the floor under staleness, and \cref{tab:source} gives 2.2 with both sources, against 2.1 and 2.6 for importance sampling with centering. On Qwen3-1.7B DLC stays within 1.2 times the floor through step 40 and ends at 0.6. In panels a and b of \cref{fig:slc}, resynchronizing importance sampling at the same threshold lowers its deviation from 115 to 2.6 times the floor. On Qwen3-1.7B each DLC run resynchronizes once in 40 steps, at steps 20, 35 and 36, when the effective sample size of the lag ratio falls below one half. Importance sampling with resynchronization reaches the same threshold two, four and five times in its three runs, and a fixed interval would have to resynchronize it every 11 steps to keep it inside the validity range.

\begin{table}[!htb]
\centering
\caption{\textbf{DLC and the baseline corrections under a stale sampler.} The deviation is the KL to on-policy training as a multiple of the noise floor at steps 10, 20, 30 and 40, and lower values are better. Test accuracy is in percent on the Countdown test problems at step 40, and higher values are better. The grey rows give the initial model $\theta_0$ and on-policy training. Among the baselines and DLC, the best value in each column is in bold and the second best is underlined, and a dash marks a correction without a run on that model.}
\label{tab:main}
\small
\setlength{\tabcolsep}{3.6pt}
\begin{tabular}{@{}l *{2}{cccc>{\hspace{2pt}}c}@{}}
\toprule
 & \multicolumn{5}{c}{\tabhead{Qwen3-0.6B}} & \multicolumn{5}{c}{\tabhead{Qwen3-1.7B}} \\
\cmidrule(lr){2-6}\cmidrule(l){7-11}
 & \multicolumn{4}{c}{\tabhead{Deviation at step}} &  & \multicolumn{4}{c}{\tabhead{Deviation at step}} &  \\
\cmidrule(lr){2-5}\cmidrule(lr){7-10}
\tabhead{Correction} & \tabhead{10} & \tabhead{20} & \tabhead{30} & \tabhead{40} & \tabhead{Accuracy} & \tabhead{10} & \tabhead{20} & \tabhead{30} & \tabhead{40} & \tabhead{Accuracy} \\
\midrule
\grouprow{11}{Reference} \\
\textcolor{gray}{Initial model $\theta_0$} & \textcolor{gray}{--} & \textcolor{gray}{--} & \textcolor{gray}{--} & \textcolor{gray}{--} & \textcolor{gray}{8.2} & \textcolor{gray}{--} & \textcolor{gray}{--} & \textcolor{gray}{--} & \textcolor{gray}{--} & \textcolor{gray}{33.4} \\
\textcolor{gray}{On-policy training} & \textcolor{gray}{1.0} & \textcolor{gray}{1.0} & \textcolor{gray}{1.0} & \textcolor{gray}{1.0} & \textcolor{gray}{46.5} & \textcolor{gray}{1.0} & \textcolor{gray}{1.0} & \textcolor{gray}{1.0} & \textcolor{gray}{1.0} & \textcolor{gray}{57.6} \\
\midrule
\grouprow{11}{Baselines} \\
Policy gradient & 2.0 & 178 & 116 & 345 & 0.2 & -- & -- & -- & -- & -- \\
PPO & 407 & 817 & 798 & 699 & 0.0 & -- & -- & -- & -- & -- \\
Score centering & 0.7 & 3.9 & 14.2 & 31.7 & 12.2 & 3.6 & 11.6 & 8.7 & 7.6 & 15.9 \\
Importance sampling & \textbf{0.3} & \underline{1.2} & 2.4 & 4.5 & 20.9 & 1.4 & 23.8 & 152 & 115 & 0.5 \\
Importance sampling with centering & \underline{0.5} & \textbf{0.9} & \underline{1.5} & \underline{2.1} & \textbf{32.0} & -- & -- & -- & -- & -- \\
Importance sampling with resynchronization & -- & -- & -- & -- & -- & \underline{0.8} & \underline{2.7} & \underline{2.3} & \underline{2.6} & \underline{36.6} \\
\midrule
\rowcolor{oursbg}
\textbf{DLC (ours)} & \textbf{0.3} & \textbf{0.9} & \textbf{1.3} & \textbf{1.9} & \underline{31.7} & \textbf{0.6} & \textbf{1.1} & \textbf{0.8} & \textbf{0.6} & \textbf{51.9} \\
\bottomrule
\end{tabular}
\end{table}

Test accuracy measures the trainer on its own responses and broadly orders the corrections as the deviation does. On Qwen3-0.6B under staleness, DLC and importance sampling with centering end at 31.7\% and 32.0\% at step 40, against 46.5\% for on-policy training and 12.2\% for score centering. Each of these checkpoints scores at or above $\theta_0$, so the gap reflects accuracy that the runs have not yet reached. On Qwen3-1.7B the three DLC models end at 51.9\% on average against 57.6\% for on-policy training, each within 7 percentage points of its on-policy counterpart in panel c. Importance sampling with resynchronization averages 36.6\%, since one of its runs collapses. \Cref{app:cost} extends the comparison to Qwen3-8B, where DLC and importance sampling with centering stay within about three times the floor while score centering reaches 89 times, and to continuous asynchrony on Qwen3-1.7B, where DLC ends at 1.2 times the floor.

\subsection{Mass off the support of the initial policy}
DLC without resynchronization keeps the clip and the centering and shows why DLC needs the resynchronization. On Qwen3-1.7B two of its three runs fall below the accuracy of $\theta_0$ by step 40 and produce answers of about 26 tokens, and the collapsed run of importance sampling with resynchronization writes responses up to the length limit. The deviation of these runs, computed on the sampler's prefixes, stays at the floor. Centering, ratio weighting and clipping act on the tokens that the sampler draws, so a shift toward tokens that the stale sampler rarely draws remains outside the reach of these three corrections. The trainer's own responses measure such a shift through their sequence KL to $\theta_0$, $\Ex_{p_\theta}\big[\sum_t \log p_\theta(y_t\mid h_t) - \log p_{\theta_0}(y_t\mid h_t)\big]$, and through the off-support share of this KL, the fraction on tokens with $p_{\theta_0}(y_t) < 10^{-3}$. At the two collapses of DLC without resynchronization the off-support share reaches about 90\%, against at most 14\% before them, as panel d of \cref{fig:slc} shows. Across the Qwen3-1.7B checkpoints the off-support share ranks the drop below $\theta_0$ at a Spearman correlation of 0.80, above the 0.74 of the deviation. On the sampler's prefixes, the next-token KL of the collapsed runs stays within the range of the runs that keep learning. The shift therefore starts on prefixes that the stale sampler does not reach. Resynchronization moves the sampler and the reference of the clip to the trainer's weights, so the trainer's responses become the sampler's responses again. Under DLC the off-support share stays at or below 11\% through step 40, and the three runs stay above the accuracy of $\theta_0$.

\section{Conclusion}

Training-inference mismatch reaches the gradient as two components with different dynamics. The numerical discrepancy of the engine stays approximately fixed during training, so removing its drift suffices. The policy lag comes from the trainer's own updates and feeds back through the residual bias of centering into later updates, and ratio weighting cuts this loop only until the effective sample size of the lag ratio falls below one half. DLC follows this account: it centers the discrepancy, clips the lag ratio and resynchronizes the sampler at that crossing, which keeps a stale sampler within twice the noise floor and prevents the collapses that the off-support share of the trainer's own responses reveals.

\section*{Reproducibility}
\Cref{app:setup} lists the training configuration, the mismatch sources and the measurement protocol, and \cref{app:split} gives the loss of DLC. The trainings run in verl with vLLM samplers, three per configuration on Qwen3-0.6B and Qwen3-1.7B, and the offline measurements use fixed checkpoints and fixed probe prefixes.

\bibliography{references}
\bibliographystyle{unsrtnat}

\newpage
\appendix

\section{Notation}\label{app:notation}
\Cref{tab:notation} collects the symbols of the main text.

\begin{table}[!htb]
\centering
\caption{\textbf{Notation.} Quantities with index $t$ belong to the sampled token $y_t$ at prefix $h_t$, and expectations $\Ex_q$ run over the sampler's responses.}
\label{tab:notation}
\small
\renewcommand{\arraystretch}{1.12}
\begin{tabular}{@{}l p{0.36\linewidth} @{\hspace{1.4em}} l p{0.36\linewidth}@{}}
\toprule
\tabhead{Symbol} & \tabhead{Meaning} & \tabhead{Symbol} & \tabhead{Meaning} \\
\midrule
$q$ & sampler distribution & $s_t$, $A_t$ & score and advantage of the sampled token \\
$p_\theta$ & trainer distribution at weights $\theta$ & $a$, $s$ & expected advantage and score as functions of the token \\
$\theta_0$ & initial weights & $\bar a$, $\bar s$ & sampler means $\Ex_q[a]$ and $\Ex_q[s]$ at a prefix \\
$\old$ & sampler's weights & $\bar A_t$ & on-policy mean $\Ex_p[a]$ at prefix $h_t$ \\
$\Delta$ & accumulated update $\theta - \theta_0$ & $H$, $\hat s_t$ & top-128 head and the head estimate of $\bar s_t$ \\
$\delta_t$ & sampled-token error $\log p_\theta - \log q$ & $\kappa$ & ratio of the tail masses \\
$\ell_t$ & policy lag $\log p_\theta - \log p_{\old}$ & $b$, $M$ & residual bias and its linear map, $b \approx -M\Delta$ \\
$e_t$ & numerical discrepancy $\log p_{\old} - \log q$ & $\eta k$ & learning rate times step count \\
$S_t$ & prefix offset $\sum_{t'<t}\delta_{t'}$ & $\rho_t$ & lag ratio $p_\theta/p_{\old} = e^{\ell_t}$ \\
$\mu$ & mean lag $\Ex_q[\ell_t]$ per token & $m_t$, $C_t$ & clip indicator and centering term of DLC \\
\bottomrule
\end{tabular}
\end{table}

\section{Experimental Details}\label{app:setup}
The trainings use verl \citep{sheng2024hybridflow} with FSDP2, BF16 compute and FP32 master weights. Each step draws 64 Countdown prompts with eight responses at temperature 1, with a maximum of 135 prompt and 377 response tokens. The advantage is the reward $\pm1$ minus the group mean without standard-deviation normalization \citep{liu2025understanding}, the loss is averaged over tokens, and one SGD step at learning rate $10^{-2}$ without momentum, warmup or weight decay follows each batch. To test bias accumulation, we repeat the stale score centering and importance sampling runs at a learning rate of $5\cdot10^{-3}$. The runs on Qwen3-1.7B use the same configuration with checkpoints every five steps.

\begin{table}[!htb]
\centering
\caption{\textbf{Deviation at step 40 on Qwen3-0.6B under the two statistics.} The ratio of means is the value of the text and of \cref{tab:source}, and the mean of ratios gives the same order of the corrections up to the swap of DLC and importance sampling with centering under the stale sampler. Dashes mark corrections without a run or without per-run values under that source.}
\label{tab:dev}
\small
\setlength{\tabcolsep}{5pt}
\begin{tabular}{@{}l *{3}{cc}@{}}
\toprule
 & \multicolumn{2}{c}{\tabhead{Quantized sampler}} & \multicolumn{2}{c}{\tabhead{Stale sampler}} & \multicolumn{2}{c}{\tabhead{Quantized and stale}} \\
\cmidrule(lr){2-3}\cmidrule(lr){4-5}\cmidrule(l){6-7}
\tabhead{Correction} & \tabhead{\begin{tabular}[b]{@{}c@{}}Ratio\\of means\end{tabular}} & \tabhead{\begin{tabular}[b]{@{}c@{}}Mean\\of ratios\end{tabular}} & \tabhead{\begin{tabular}[b]{@{}c@{}}Ratio\\of means\end{tabular}} & \tabhead{\begin{tabular}[b]{@{}c@{}}Mean\\of ratios\end{tabular}} & \tabhead{\begin{tabular}[b]{@{}c@{}}Ratio\\of means\end{tabular}} & \tabhead{\begin{tabular}[b]{@{}c@{}}Mean\\of ratios\end{tabular}} \\
\midrule
Policy gradient & 12.7 & 16.4 & 345 & 516 & 107 & -- \\
PPO w/o clip & 336 & 429 & 534 & 718 & -- & -- \\
PPO & 14.3 & 19.5 & 699 & 804 & -- & -- \\
Importance sampling & 0.74 & 1.09 & 4.50 & 6.31 & 3.91 & 4.79 \\
Score centering & 0.80 & 1.22 & 31.7 & 45.1 & 28.1 & 35.7 \\
Importance sampling with centering & 1.11 & 1.53 & 2.10 & 2.51 & 2.61 & 3.64 \\
\midrule
\rowcolor{oursbg}
\textbf{DLC (ours)} & -- & -- & 1.93 & 2.70 & 2.17 & 2.76 \\
\bottomrule
\end{tabular}
\end{table}

The probe prefixes are the prefixes of responses from $\theta_0$ to a fixed set of prompts. The deviation is the full-vocabulary KL from the trained model to the paired on-policy model, averaged over these prefixes. The residual bias $b$ is the corrected gradient minus the on-policy gradient on the same regenerated batch, and the gradient noise is the difference of two independent on-policy gradients. The off-support share is the fraction of the sequence KL of the trainer's own responses to $\theta_0$ on tokens with $p_{\theta_0}(y_t) < 10^{-3}$, computed on 256 responses of the run to the test prompts. \Cref{tab:dev} reports the deviation both as the ratio of the mean KL to the mean floor, used in the text, and as the mean of the per-run ratios.

\section{Step Cost, a Larger Model and Continuous Asynchrony}\label{app:cost}
\Cref{tab:cost} splits the wall-clock time of a training step on Qwen3-1.7B under a stale sampler into generation, the forward pass at the sampler's weights and the parameter update, averaged over the three runs of each correction. DLC adds two costs to the step. The sampler reports its top-128 head, which doubles the generation time as it does for score centering, and the trainer computes the head log-probabilities at the sampler's weights in a forward pass that PPO with the sampler's weights as reference also needs. In our implementation about 19 of the 54 seconds of this forward come from a second reference forward on the KL path of verl, whose output does not enter the loss at a zero KL coefficient. The resynchronization copies the trainer's weights to the sampler once per run in 40 steps and takes at most 1.5 seconds per copy, against about 4500 seconds of training per run. On-policy training copies the weights after each step and spends 1.2 seconds per step on the copy.

\begin{table}[!htb]
\centering
\caption{\textbf{Wall-clock time per training step on Qwen3-1.7B under a stale sampler.} Times are seconds per step averaged over 40 steps and three runs, and the step time also contains the advantage computation and the bookkeeping of verl. The last column counts the weight copies to the sampler in 40 steps.}
\label{tab:cost}
\small
\setlength{\tabcolsep}{5pt}
\begin{tabular}{@{}l ccccc@{}}
\toprule
\tabhead{Correction} & \tabhead{Generation} & \tabhead{\begin{tabular}[b]{@{}c@{}}Forward at the\\sampler's weights\end{tabular}} & \tabhead{Update} & \tabhead{Step} & \tabhead{\begin{tabular}[b]{@{}c@{}}Copies in\\40 steps\end{tabular}} \\
\midrule
\grouprow{6}{Reference} \\
\textcolor{gray}{On-policy training} & \textcolor{gray}{6.5} & \textcolor{gray}{0.1} & \textcolor{gray}{37.1} & \textcolor{gray}{50.4} & \textcolor{gray}{40} \\
\midrule
\grouprow{6}{Corrections} \\
Score centering & 14.7 & 0.1 & 42.1 & 62.2 & 0 \\
Importance sampling & 6.7 & 0.1 & 40.8 & 53.2 & 0 \\
Importance sampling with resynchronization & 7.0 & 0.1 & 39.5 & 52.7 & 3.7 \\
DLC w/o resynchronization & 14.5 & 56.6 & 41.7 & 118.2 & 0 \\
\midrule
\rowcolor{oursbg}
\textbf{DLC (ours)} & 13.7 & 54.2 & 40.1 & 113.2 & 1 \\
\bottomrule
\end{tabular}
\end{table}

On Qwen3-8B, one run per correction trains under the stale sampler of the main text, and \cref{tab:8b} gives the deviation and the test accuracy at step 40. DLC and importance sampling with centering stay within about three times the noise floor, at 3.1 and 2.9, while policy gradient and M2PO end at zero accuracy and PPO at 20.7\%. Score centering reaches 89 times the floor, so on the larger model the policy lag again leaves the reach of centering. DLC ends at 49.6\% test accuracy, against 51.8\% for importance sampling with centering and 55.3\% for on-policy training.

\begin{table}[!htb]
\centering
\caption{\textbf{Corrections on Qwen3-8B under a stale sampler at step 40.} The deviation is the KL to on-policy training as a multiple of the noise floor, and the accuracy is the test accuracy in percent. Each row is one run.}
\label{tab:8b}
\small
\setlength{\tabcolsep}{6pt}
\begin{tabular}{@{}l cc@{}}
\toprule
\tabhead{Correction} & \tabhead{Deviation} & \tabhead{Accuracy} \\
\midrule
\grouprow{3}{Reference} \\
\textcolor{gray}{Initial model $\theta_0$} & \textcolor{gray}{--} & \textcolor{gray}{35.4} \\
\textcolor{gray}{On-policy training} & \textcolor{gray}{1.0} & \textcolor{gray}{55.3} \\
\midrule
\grouprow{3}{Baselines} \\
Policy gradient & 6470 & 0.0 \\
PPO & 1100 & 20.7 \\
Score centering & 89.1 & \underline{51.4} \\
Importance sampling & 6.2 & 51.0 \\
Importance sampling with centering & \textbf{2.9} & \textbf{51.8} \\
Importance sampling with resynchronization & 5.2 & 50.4 \\
M2PO & 10768 & 0.0 \\
VCPO & 14.0 & 42.6 \\
\midrule
\rowcolor{oursbg}
\textbf{DLC (ours)} & \underline{3.1} & 49.6 \\
\bottomrule
\end{tabular}
\end{table}

In continuous asynchronous training the sampler receives the trainer's weights after each step, so each token carries the version that generated it. With a maximum version gap of four on Qwen3-1.7B, the sampled tokens lag the trainer by 2.85 versions on average, and DLC clips the ratio against the generating version of each token. The admission variant defers a response whose effective sample size of the lag ratio falls below one half and regenerates it with the newer weights. \Cref{tab:async} gives one run per correction at step 40 together with the time per step. DLC ends at 1.2 times the floor and 55.9\% test accuracy, and the admission variant ends at 0.8 times the floor after deferring 0.3\% of the responses, 68 in 40 steps.

\begin{table}[!htb]
\centering
\caption{\textbf{Corrections on Qwen3-1.7B under continuous asynchrony at step 40.} The sampler receives the weights after each step with a maximum version gap of four. The deviation and the accuracy are as in \cref{tab:8b}, and the step time is the mean wall-clock time per step in seconds. Each row is one run.}
\label{tab:async}
\small
\setlength{\tabcolsep}{6pt}
\begin{tabular}{@{}l ccc@{}}
\toprule
\tabhead{Correction} & \tabhead{Deviation} & \tabhead{Accuracy} & \tabhead{Step time} \\
\midrule
\grouprow{4}{Reference} \\
\textcolor{gray}{Initial model $\theta_0$} & \textcolor{gray}{--} & \textcolor{gray}{33.4} & \textcolor{gray}{--} \\
\textcolor{gray}{On-policy training} & \textcolor{gray}{1.0} & \textcolor{gray}{56.8} & \textcolor{gray}{--} \\
\midrule
\grouprow{4}{Baselines} \\
Policy gradient & 188 & 0.0 & 51.8 \\
Importance sampling & 1.4 & 52.9 & 43.7 \\
M2PO & 2.4 & 23.8 & 78.3 \\
\midrule
\rowcolor{oursbg}
\textbf{DLC (ours)} & \underline{1.2} & \textbf{55.9} & 80.1 \\
\rowcolor{oursbg}
\textbf{DLC with admission (ours)} & \textbf{0.8} & \underline{55.3} & 94.2 \\
\bottomrule
\end{tabular}
\end{table}

\section{Drift and the Corrections at a Prefix}\label{app:prefix}
This section fixes one prefix and writes sums over the vocabulary $V$, with $p_v$, $q_v$, $s_v$ and $a_v$ as in \cref{sec:setup}. The score has mean zero on policy, since the probabilities sum to one,
\begin{equation}
\Ex_p[s] = \sum_{v\in V} p_v\,\nabla_\theta \log p_v = \sum_{v\in V} \nabla_\theta\, p_v = \nabla_\theta \sum_{v\in V} p_v = 0 .
\label{eq:app-zero}
\end{equation}
The on-policy update at the prefix is therefore $\Ex_p[a\,s] = \Ex_p\big[(a - \Ex_p[a])\,s\big] = \mathrm{Cov}_p(a, s)$, and a constant shift of the advantage leaves it unchanged. The sampler's probabilities do not depend on $\theta$, so the gradient of the KL from the sampler to the trainer is
\begin{equation}
\nabla_\theta\,\mathrm{KL}(q\,\|\,p_\theta) = \nabla_\theta \sum_{v} q_v\big(\log q_v - \log p_v\big) = -\sum_v q_v\,s_v = -\Ex_q[s],
\end{equation}
which together with \cref{eq:app-zero} gives the second line of \cref{eq:drift}. The first line of \cref{eq:drift} is the definition of the covariance. Under the sampler a constant shift $c$ of the advantage adds $c\,\bar s$ to the uncorrected update, whereas the centered update stays unchanged because $\Ex_q[s - \bar s] = 0$. Score centering thus gives the sampler's update the invariance to a constant advantage that the on-policy update has.

The untruncated ratio reweights the sampler's tokens to the trainer's, $\Ex_q[(p/q)\,a\,s] = \sum_v p_v\,a_v\,s_v = \Ex_p[a\,s]$ on tokens with $q_v > 0$, so it removes the drift together with the rest of the bias at the prefix. The truncated weight $w_v = \min(p_v/q_v, 2)$ falls short of the ratio by $p_v/q_v - w_v = \max(p_v/q_v - 2, 0)$, which gives
\begin{equation}
\begin{aligned}
\Ex_q[w\,a\,s] &= \sum_v q_v\Big(\frac{p_v}{q_v} - \max\Big(\frac{p_v}{q_v} - 2,\, 0\Big)\Big)\,a_v\,s_v \\
&= \Ex_p[a\,s] - \sum_v q_v\,\max\Big(\frac{p_v}{q_v} - 2,\, 0\Big)\,a_v\,s_v \\
&= \Ex_p[a\,s] - \bar a\sum_v q_v\,\max\Big(\frac{p_v}{q_v} - 2,\, 0\Big)\,s_v - \sum_v q_v\,\max\Big(\frac{p_v}{q_v} - 2,\, 0\Big)\,(a_v - \bar a)\,s_v .
\end{aligned}
\label{eq:app-trunc}
\end{equation}
The second term of the last line depends on the advantages only through their mean $\bar a$, so the truncation brings back a drift on the truncated tokens, and the third term is a covariance on the same tokens. Both terms vanish when the ratio stays below 2 on the tokens of the prefix.

The sampler reports $q$ on its head $H$ only. The tail of the estimate takes $\kappa\,p_v$ for $v \notin H$, and $\kappa$ matches the tail mass of the sampler, $\sum_{v\notin H}\kappa\,p_v = \kappa\,\big(1 - \sum_{v\in H} p_v\big) = 1 - \sum_{v\in H} q_v$. With this tail the estimate of the mean score follows in two steps,
\begin{equation}
\hat s = \sum_{v\in H} q_v\,s_v + \kappa\sum_{v\notin H} p_v\,s_v = \sum_{v\in H} q_v\,s_v - \kappa\sum_{v\in H} p_v\,s_v = \sum_{v\in H}\big(q_v - \kappa\,p_v\big)\,s_v,
\end{equation}
where the second step replaces $\sum_{v\notin H} p_v\,s_v$ by $-\sum_{v\in H} p_v\,s_v$ through \cref{eq:app-zero}. The estimate therefore needs the trainer's scores on the head alone. Its error is
\begin{equation}
\bar s - \hat s = \sum_{v\notin H} q_v\,s_v - \kappa\sum_{v\notin H} p_v\,s_v = \sum_{v\notin H}\big(q_v - \kappa\,p_v\big)\,s_v ,
\end{equation}
which vanishes when the sampler's tail is proportional to the trainer's tail and which enters \cref{eq:exact} as the tail term.

\section{Exact Split of the Sampled-Token Error}\label{app:parts}
The split of \cref{eq:split} is the logarithm of the factorization of the importance ratio at the sampler's weights,
\begin{equation}
\frac{p_\theta(y_t\mid h_t)}{q(y_t\mid h_t)} = \frac{p_\theta(y_t\mid h_t)}{p_{\old}(y_t\mid h_t)}\cdot\frac{p_{\old}(y_t\mid h_t)}{q(y_t\mid h_t)} = e^{\ell_t}\,e^{e_t}.
\end{equation}
Both probabilities in $e_t$ are evaluated at the sampler's weights $\old$, so $e_t$ depends on $\old$, on the engine and on the token, and between two synchronizations it stays the same for a given token while the trainer moves. The lag depends on the trainer only through $p_\theta$. With the version gap $\Delta_{\old} = \theta - \old$ and the score $s^{\old}_t$ at the sampler's weights, a Taylor expansion of $\log p_\theta$ around $\old$ gives
\begin{equation}
\ell_t = \Delta_{\old}^\top s^{\old}_t + \tfrac12\,\Delta_{\old}^\top\,\nabla^2_\theta \log p_{\old}(y_t\mid h_t)\,\Delta_{\old} + O\big(\|\Delta_{\old}\|^3\big).
\label{eq:app-lag}
\end{equation}
Under staleness the sampler keeps $\old = \theta_0$, so $\Delta_{\old} = \Delta$. The scores at $\theta$ and at $\theta_0$ differ by a term of first order in $\Delta$, so replacing $s^{\old}_t$ by $s_t$ in the first term changes $\ell_t$ at second order, which gives $\ell_t \approx \Delta^\top s_t$ of \cref{sec:sources}.

With $e_t = 0$ the sampled tokens follow $p_{\old}$, and at a prefix the mean lag is
\begin{equation}
\mu = \Ex_{p_{\old}}[\ell] = \sum_v p_{\old}(v)\big(\log p_\theta(v) - \log p_{\old}(v)\big) = -\mathrm{KL}\big(p_{\old}\,\|\,p_\theta\big).
\end{equation}
The first term of \cref{eq:app-lag} has mean $\Delta_{\old}^\top\,\Ex_{p_{\old}}[s^{\old}] = 0$ by \cref{eq:app-zero} at $\old$. The second term has mean $-\tfrac12\Delta_{\old}^\top F_{\old}\Delta_{\old}$, since the expected Hessian of the log-probability is minus the Fisher information $F_{\old} = \Ex_{p_{\old}}\big[s^{\old}(s^{\old})^\top\big]$. The mean lag is thus quadratic in the version gap,
\begin{equation}
\mu = -\tfrac12\,\Delta_{\old}^\top F_{\old}\,\Delta_{\old} + O\big(\|\Delta_{\old}\|^3\big), \qquad
\mathrm{Var}_{p_{\old}}(\ell) = \Delta_{\old}^\top F_{\old}\,\Delta_{\old} + O\big(\|\Delta_{\old}\|^3\big),
\label{eq:app-moments}
\end{equation}
where the variance comes from the first term of \cref{eq:app-lag}. Panel a of \cref{fig:loop} measures the ratio of $|\mu|$ to $\|\Delta\|^2$, which by \cref{eq:app-moments} equals half the Fisher curvature along the direction of $\Delta$. The two moments also satisfy $\mathrm{Var}(\ell) = -2\mu$ to second order, which is the relation that a Gaussian lag with a mean-one ratio satisfies exactly, as \cref{app:ess} shows. Over a response the chain rule of the KL gives $\Ex_q\big[\sum_t \ell_t\big] = -\mathrm{KL}\big(p_{\old}(y)\,\|\,p_\theta(y)\big)$ for the sequence distributions, so the stale sequence offset decreases as the sequence KL from the sampler's weights to the trainer increases.

\section{Residual Bias of the Corrections}\label{app:derivation}
This section derives each term of \cref{eq:exact} from a change of measure and then the first-order forms of \cref{eq:bias}. Throughout, $\delta_v = \log p_v - \log q_v$ at a prefix, so that $q_v = p_v\,e^{-\delta_v}$, and the sampled-token error is $\delta_t$.

The prefix $h_t$ consists of the tokens $y_{<t}$, and its probability under a policy is the product of the next-token probabilities. The density of the trainer's prefix distribution with respect to the sampler's is therefore
\begin{equation}
\frac{\mu^p_t(h_t)}{\mu^q_t(h_t)} = \prod_{j<t}\frac{p(y_j\mid h_j)}{q(y_j\mid h_j)} = \exp\Big(\sum_{j<t}\delta_j\Big) = e^{S_t},
\end{equation}
and the prefix term of \cref{eq:exact} becomes an expectation under the sampler,
\begin{equation}
\big(\Ex_{\mu^q_t} - \Ex_{\mu^p_t}\big)\,\mathrm{Cov}_q(a^q_t, s_t) = \Ex_{\mu^q_t}\Big[\big(1 - e^{S_t}\big)\,\mathrm{Cov}_q(a^q_t, s_t)\Big].
\label{eq:app-prefix}
\end{equation}
The continuation term changes the measure after the token. The expected advantage $a^\pi_t(v)$ averages the return over continuations from $\pi$, and the same change of measure on the tokens after $t$ gives
\begin{equation}
a^p_t(v) = \Ex_q\Big[A\prod_{j>t} e^{\delta_j}\,\Big|\, h_t,\, y_t = v\Big], \qquad
a^q_t(v) - a^p_t(v) = \Ex_q\Big[A\Big(1 - \prod_{j>t} e^{\delta_j}\Big)\,\Big|\, h_t,\, y_t = v\Big].
\label{eq:app-cont}
\end{equation}
The continuation term is the covariance of this difference with the score at the prefix. The token term compares two covariances of the same function $a = a^p_t$ under $q$ and under $p$. With $\bar a^q = \Ex_q[a]$, $\bar a^p = \Ex_p[a]$ and $\bar s = \Ex_q[s]$, it follows in four steps,
\begin{equation}
\begin{aligned}
\mathrm{Cov}_q(a, s) - \mathrm{Cov}_p(a, s) &= \sum_v q_v\,a_v\,s_v - \bar a^q\,\bar s - \sum_v p_v\,a_v\,s_v \\
&= \sum_v p_v\big(e^{-\delta_v} - 1\big)\,a_v\,s_v - \bar a^q\,\bar s \\
&= \sum_v p_v\big(e^{-\delta_v} - 1\big)\big(a_v - \bar a^p\big)\,s_v + \bar a^p\,\bar s - \bar a^q\,\bar s \\
&= \Ex_p\Big[\big(e^{-\delta} - 1\big)\big(a - \bar a^p\big)\,s\Big] - \big(\bar a^q - \bar a^p\big)\,\bar s,
\end{aligned}
\label{eq:app-token}
\end{equation}
where the first step uses $\mathrm{Cov}_p(a, s) = \Ex_p[a\,s]$ from \cref{eq:app-zero}, the second step substitutes $q_v = p_v e^{-\delta_v}$, and the third step adds and subtracts $\bar a^p\sum_v p_v(e^{-\delta_v} - 1)\,s_v = \bar a^p\,\bar s$. The token term vanishes when the token does not change the expected advantage, $a_v = \bar a^p$ for each $v$, or when the error is zero. The tail term is $\Ex_{\mu^q_t}\big[\Ex_q[a^q_t]\,(\bar s_t - \hat s_t)\big]$ with the tail error of \cref{app:prefix}. Constant baselines in the advantage cancel from the centered update under $q$ and from the on-policy update under $p$, so they do not enter the four terms.

The prefix, continuation and token terms are each of first order in the error $\delta$. Expanding $e^{S_t} = 1 + S_t + O(S_t^2)$, $e^{-\delta} - 1 = -\delta + O(\delta^2)$ and $\prod_{j>t} e^{\delta_j} = 1 + \sum_{j>t}\delta_j + O(\delta^2)$, and noting that $\bar a^q - \bar a^p$ and $\bar s$ are each of first order so that their product is of second order, turns the prefix term, the continuation term and the token term, in this order, into
\begin{equation}
\begin{aligned}
&-\Ex_{\mu^q_t}\big[S_t\,\mathrm{Cov}_q(a^q_t, s_t)\big], \\
&-\Ex_{\mu^p_t}\,\mathrm{Cov}_q\Big(\Ex_q\Big[A\sum_{j>t}\delta_j\,\Big|\,h_t,\, y_t\Big],\ s_t\Big), \\
&-\Ex_{\mu^p_t}\,\Ex_p\big[(a^p_t - \bar a^p_t)\,\delta\,s\big].
\end{aligned}
\label{eq:app-first}
\end{equation}
On a sampled token the advantage $A_t$ is a one-sample estimate of $a$, and the centered score $s_t - \bar s_t$ equals $s_t$ up to a first-order term. The product of $S_t$ with $\bar s_t$ is therefore of second order, and the prefix term samples as $-A_t S_t s_t$. The token term samples as $-(A_t - \bar A_t)\,\delta_t\,s_t$ with $\bar A_t = \Ex_p[a]$ at $h_t$, since replacing the expectation under $p$ by the sampled token from $q$ changes it at second order. Under staleness $\delta_t = \ell_t$, and the sum of the two sampled forms gives the first line of \cref{eq:bias}. The prediction of \cref{sec:loop} leaves out the continuation term, which needs the errors after each token and an estimate of their covariance with the return.

Importance sampling weights each sampled token by $w_t$ and leaves the prefix distribution of the sampler, so its bias splits as \cref{eq:exact} does, with $\mathrm{Cov}_q(a^q_t, s_t)$ replaced by $\Ex_q[w\,a^q_t\,s_t]$ and without a tail term, since it does not use a head estimate. By \cref{eq:app-trunc} the token term of importance sampling is exactly
\begin{equation}
\Ex_{\mu^p_t}\Big[\Ex_q[w\,a^p_t\,s_t] - \Ex_p[a^p_t\,s_t]\Big] = -\Ex_{\mu^p_t}\Big[\sum_v q_v\,\max\big(e^{\delta_v} - 2,\, 0\big)\,a^p_t(v)\,s_v\Big],
\end{equation}
which vanishes on prefixes whose ratios stay below 2. The prefix term of importance sampling agrees with that of score centering at first order, since $w = 1 + O(\delta)$. The sum of the prefix and token terms of importance sampling samples as $-A_t\big(S_t + \max(e^{\ell_t} - 2, 0)\big)\,s_t$ under staleness, with the truncation weight kept exact, which gives the second line of \cref{eq:bias}.

Under staleness the lag and the prefix offset are linear in $\Delta$ to first order, $\ell_t \approx s_t^\top\Delta$ and $S_t = \sum_{t'<t}\ell_{t'} \approx \sum_{t'<t} s_{t'}^\top\Delta$. Substituting both into the sampled forms, with the truncation weight of importance sampling at zero inside the validity range, gives
\begin{equation}
\begin{aligned}
(A_t - \bar A_t)\,\ell_t\,s_t + A_t\,S_t\,s_t &\approx s_t\Big((A_t - \bar A_t)\,s_t + A_t\sum_{t'<t} s_{t'}\Big)^{\!\top}\Delta, \\
A_t\,S_t\,s_t &\approx A_t\,s_t\sum_{t'<t} s_{t'}^\top\,\Delta,
\end{aligned}
\end{equation}
and taking the expectation and the sum over $t$ gives $M_\mathrm{c}$ and $M_\mathrm{w}$ of \cref{eq:bias}. The two maps share the prefix part $\sum_t\Ex\big[A_t\,s_t\sum_{t'<t} s_{t'}^\top\big]$, and $M_\mathrm{c}$ adds the token part
\begin{equation}
M_\mathrm{c} - M_\mathrm{w} = \sum_t \Ex\Big[(A_t - \bar A_t)\,s_t\,s_t^\top\Big],
\label{eq:app-mdiff}
\end{equation}
a symmetric matrix whose quadratic form $u^\top(M_\mathrm{c} - M_\mathrm{w})u = \sum_t\Ex\big[(A_t - \bar A_t)(s_t^\top u)^2\big]$ is negative along directions that the scores of tokens with an advantage below their prefix mean dominate.

\section{Exact-Weight Recomputation of the Residual Bias}\label{app:exact}
The prediction of $b$ for the residual-bias link in \cref{sec:loop} uses the per-token weights of \cref{eq:bias}. The exact recomputation replaces them by $A_t(1 - e^{S_t})$ for the prefix term, $A_t(1 - e^{\ell_t})$ for the score centering token term and $-A_t\max(e^{\ell_t} - 2, 0)$ for the importance sampling truncation, on the same checkpoints, batches and backward passes. In both the linearized and the exact token weights of score centering, the sampled advantage $A_t$ takes the place of $A_t - \bar A_t$, since $\bar A_t$ requires continuations from each prefix.

\begin{table}[!htb]
\centering
\caption{\textbf{Predicted against measured residual bias with linearized and exact weights.} The cosine to the measured bias and the norm ratio $\|b_\mathrm{pred}\|/\|b\|$ are means over three runs. The exact weights reduce the norm ratio from up to 3400 to between 1.9 and 7.9, and the cosine stays above the gradient noise of at most 0.07.}
\label{tab:exact}
\small
\setlength{\tabcolsep}{6pt}
\begin{tabular}{@{}ll cc cc@{}}
\toprule
 & & \multicolumn{2}{c}{\tabhead{Cosine at step}} & \multicolumn{2}{c}{\tabhead{Norm ratio at step}} \\
\cmidrule(lr){3-4}\cmidrule(l){5-6}
\tabhead{Correction} & \tabhead{Weights} & \tabhead{20} & \tabhead{40} & \tabhead{20} & \tabhead{40} \\
\midrule
\grouprow{6}{Stale sampler} \\
\multirow{2}{*}{Score centering} & linearized & 0.55 & 0.16 & 136 & 3399 \\
 & exact & 0.28 & 0.09 & 4.1 & 6.1 \\
\addlinespace[2pt]
\multirow{2}{*}{Importance sampling} & linearized & 0.41 & 0.44 & 23 & 93 \\
 & exact & 0.15 & 0.15 & 7.7 & 7.9 \\
\midrule
\grouprow{6}{Quantized and stale} \\
\multirow{2}{*}{Score centering} & linearized & 0.30 & 0.09 & 129 & 2734 \\
 & exact & 0.22 & 0.15 & 2.8 & 4.3 \\
\addlinespace[2pt]
\multirow{2}{*}{Importance sampling} & linearized & 0.16 & 0.34 & 14 & 43 \\
 & exact & 0.13 & 0.14 & 1.9 & 3.2 \\
\bottomrule
\end{tabular}
\end{table}

\section{Unrolling the Loop}\label{app:loop}
This section solves \cref{eq:loop} within one synchronization period. The sampler resynchronizes at step zero, so $\Delta_0 = 0$, and the derivation treats $M$ as constant over the period and writes the on-policy gradient as $g_k = \bar g + \xi_k$ with a mean $\bar g$ and zero-mean noise $\xi_k$.

Applying \cref{eq:loop} $k$ times from $\Delta_0 = 0$ gives
\begin{equation}
\Delta_k = \eta\sum_{j=0}^{k-1}(I - \eta M)^{k-1-j}\,g_j, \qquad
\Ex[\Delta_k] = \eta\sum_{j=0}^{k-1}(I - \eta M)^{j}\,\bar g .
\label{eq:app-unroll}
\end{equation}
For a left eigenvector $u$ of $M$ with eigenvalue $\lambda$, so that $u^\top(I - \eta M)^j = (1 - \eta\lambda)^j\,u^\top$, the geometric sum has the closed form
\begin{equation}
u^\top\Ex[\Delta_k] = \eta\,\frac{1 - (1 - \eta\lambda)^k}{\eta\lambda}\,u^\top\bar g = \frac{1 - (1 - \eta\lambda)^k}{\lambda}\,u^\top\bar g \;\longrightarrow\; \frac{1 - e^{-\lambda\,\eta k}}{\lambda}\,u^\top\bar g ,
\label{eq:app-eigen}
\end{equation}
where the limit takes $\eta \to 0$ at a fixed product $\eta k$, so that $(1 - \eta\lambda)^k = \exp\big(k\log(1 - \eta\lambda)\big) \to e^{-\lambda\eta k}$. In this limit the update depends on the learning rate and the step count only through $\eta k$. For $\lambda > 0$ the component saturates at $u^\top\bar g/\lambda$, and for $\lambda < 0$ it grows as $(e^{|\lambda|\eta k} - 1)/|\lambda|$, faster than the on-policy component $\eta k\,u^\top\bar g$.

Expanding $(I - \eta M)^j = I - j\eta M + O(j^2\eta^2\|M\|^2)$ inside \cref{eq:app-unroll} and summing over $j$ gives
\begin{equation}
\Ex[\Delta_k] = \eta k\,\bar g - \eta^2\frac{k(k-1)}{2}\,M\bar g + O\big((\eta k)^3\|M\|^2\|\bar g\|\big)
= \eta k\,\bar g - \tfrac12(\eta k)^2\,M\bar g + \tfrac12\,\eta\,(\eta k)\,M\bar g + O\big((\eta k)^3\big).
\end{equation}
The third term vanishes as $\eta \to 0$ at fixed $\eta k$, which recovers the second-order form of \cref{sec:loop}. For two vectors $x \neq 0$ and $y$ with $\|y\| \ll \|x\|$, the norm expands as $\|x + y\| = \|x\| + x^\top y/\|x\| + O(\|y\|^2/\|x\|)$. With $x = \eta k\,\bar g$ and $y = -\tfrac12(\eta k)^2 M\bar g$, the norm of the mean update is
\begin{equation}
\big\|\Ex[\Delta_k]\big\| = \eta k\,\|\bar g\| - \tfrac12(\eta k)^2\,\frac{\bar g^\top M\,\bar g}{\|\bar g\|} + O\big((\eta k)^3\big).
\end{equation}
Score centering and importance sampling share $\bar g$ and differ in $M$, so the excess update norm of score centering over importance sampling is
\begin{equation}
E = \tfrac12(\eta k)^2\,\frac{\bar g^\top\big(M_\mathrm{w} - M_\mathrm{c}\big)\,\bar g}{\|\bar g\|} + O\big((\eta k)^3\big)
= -\tfrac12(\eta k)^2\,\frac{1}{\|\bar g\|}\sum_t\Ex\Big[(A_t - \bar A_t)\,\big(s_t^\top\bar g\big)^2\Big] + O\big((\eta k)^3\big),
\label{eq:app-excess}
\end{equation}
where the second step uses \cref{eq:app-mdiff}. The quadratic coefficient is positive when the tokens with an advantage below their prefix mean carry more of the squared projection of the score on $\bar g$, which panel b of \cref{fig:loop} measures as the larger lag of these tokens under score centering. Biases $b^0$ that do not depend on $\Delta$ instead add $\eta k\,b^0$ to $\Ex[\Delta_k]$ and change the norm by $\eta k\,(\|\bar g + b^0\| - \|\bar g\|)$, a term linear in $\eta k$. The fit $E = a\,\eta k + c\,(\eta k)^2$ of \cref{sec:loop} therefore separates the biases that are fixed in $\Delta$, in the linear coefficient $a$, from the loop, in the quadratic coefficient $c$.

The bias $b_k$ moves the trainer by $\eta\,b_k$ in step $k$, and by the first-order lag $\ell_t \approx s_t^\top\Delta$ it changes the lag of token $t$ by $\eta\,s_t^\top b_k$. The token part $-\Ex_{t'}\big[(A_{t'} - \bar A_{t'})\,\ell_{t'}\,s_{t'}\big]$ of the bias of score centering contributes the change
\begin{equation}
-\eta\,\Ex_{t'}\Big[\big(s_t^\top s_{t'}\big)\big(A_{t'} - \bar A_{t'}\big)\,\ell_{t'}\Big],
\end{equation}
where the average runs over the $N$ tokens of the batch. The term $t' = t$ contributes $-\tfrac{\eta}{N}\,\|s_t\|^2\,(A_t - \bar A_t)\,\ell_t$, which has the sign of $\ell_t$ when $A_t < \bar A_t$. The lag of a token whose advantage lies below its prefix mean therefore grows in magnitude by the factor $1 + \tfrac{\eta}{N}\|s_t\|^2\,|A_t - \bar A_t|$ per step from its own contribution, and the tokens with aligned scores add to this growth through the terms $t' \neq t$.

\section{Effective Sample Size and Truncation under a Gaussian Lag}\label{app:ess}
With $e_t = 0$ the lag ratio has mean one under the sampler at each prefix,
\begin{equation}
\Ex_{p_{\old}}[\rho] = \sum_v p_{\old}(v)\,\frac{p_\theta(v)}{p_{\old}(v)} = \sum_v p_\theta(v) = 1,
\end{equation}
and its second moment is one plus the chi-squared divergence,
\begin{equation}
\begin{aligned}
\Ex_{p_{\old}}[\rho^2] &= \sum_v \frac{p_\theta(v)^2}{p_{\old}(v)} = \sum_v \frac{\big(p_\theta(v) - p_{\old}(v)\big)^2}{p_{\old}(v)} + 2\sum_v p_\theta(v) - \sum_v p_{\old}(v) \\
&= 1 + \chi^2\big(p_\theta\,\|\,p_{\old}\big).
\end{aligned}
\end{equation}
For weights $\rho_1, \dots, \rho_N$ the Kish effective sample size as a fraction of the batch is $\big(\sum_i\rho_i\big)^2/\big(N\sum_i\rho_i^2\big)$, which tends to $\Ex[\rho]^2/\Ex[\rho^2] = 1/(1 + \chi^2)$ for large $N$. To second order $\chi^2(p_\theta\|p_{\old}) = \Ex_{p_{\old}}\big[(\rho - 1)^2\big] = \Ex_{p_{\old}}[\ell^2] + O(\|\Delta_{\old}\|^3) = \Delta_{\old}^\top F_{\old}\Delta_{\old} + O(\|\Delta_{\old}\|^3)$, so the threshold of one half on the effective sample size bounds the version gap in the Fisher metric of the sampler's weights.

For a Gaussian lag $\ell \sim \mathcal N(\mu, \sigma^2)$ the moments of the ratio follow from the moment generating function $\Ex[e^{r\ell}] = e^{r\mu + r^2\sigma^2/2}$. The constraint $\Ex[\rho] = 1$ at $r = 1$ fixes $\mu = -\sigma^2/2$, and at $r = 2$
\begin{equation}
\Ex[\rho^2] = e^{2\mu + 2\sigma^2} = e^{2\mu - 4\mu} = e^{-2\mu}, \qquad \frac{1}{\Ex[\rho^2]} = e^{2\mu}.
\end{equation}
The Gaussian effective sample size therefore falls below one half at $\mu = -\tfrac12\log 2 \approx -0.35$ nats per token. The share of tokens with a ratio below one half is
\begin{equation}
\Pr(\rho < \tfrac12) = \Pr(\ell < -\log 2) = \Phi\Big(\frac{-\log 2 - \mu}{\sqrt{-2\mu}}\Big),
\end{equation}
with $\Phi$ the standard normal distribution function, and panel a of \cref{fig:range} compares this value with the measured share.

Without the Gaussian assumption, the cumulant generating function $K(r) = \log\Ex[e^{r\ell}] = \sum_{n\geq1}\nu_n\,r^n/n!$ with cumulants $\nu_1 = \mu$, $\nu_2 = \sigma^2$, $\nu_3$ and $\nu_4$ satisfies $K(1) = 0$ by the mean-one constraint. Subtracting $4K(1)$ from $K(2)$ removes the variance,
\begin{equation}
\log\Ex[\rho^2] = K(2) - 4K(1) = \big(2 - 4\big)\nu_1 + \big(2 - 2\big)\nu_2 + \Big(\tfrac{4}{3} - \tfrac{2}{3}\Big)\nu_3 + \Big(\tfrac{2}{3} - \tfrac{1}{6}\Big)\nu_4 + \dots
= -2\mu + \tfrac23\,\nu_3 + \tfrac12\,\nu_4 + \dots,
\end{equation}
so the effective sample size is $e^{2\mu - \frac23\nu_3 - \frac12\nu_4 - \dots}$. Positive third or fourth cumulants of the lag lower the effective sample size below the Gaussian value $e^{2\mu}$ at the same mean lag, so a lag that concentrates on few tokens with large ratios crosses one half at a smaller $|\mu|$. On Qwen3-1.7B the crossing at a mean lag of $-0.15$ nats, against the Gaussian $-0.35$, gives $\tfrac23\nu_3 + \tfrac12\nu_4 + \dots \approx \log 2 - 0.30 \approx 0.39$ at step 11.

The truncation of importance sampling at 2 keeps the excess $\max(\rho - 2, 0)$ out of the weight. For the Gaussian lag with $\mu = -\sigma^2/2$, the lognormal call formula gives the expected excess
\begin{equation}
\Ex\big[\max(\rho - 2,\, 0)\big] = \Phi\Big(\frac{\sigma^2/2 - \log 2}{\sigma}\Big) - 2\,\Phi\Big(\frac{-\sigma^2/2 - \log 2}{\sigma}\Big), \qquad \sigma^2 = -2\mu,
\end{equation}
which is the mean of the truncation weight in \cref{eq:bias} and which increases with $|\mu|$. The clip of DLC drops a token with positive advantage when $\rho > 1 + \varepsilon$ and a token with negative advantage when $\rho < 1 - \varepsilon$ or $\rho > c$. Under the Gaussian lag the two shares are
\begin{equation}
\begin{aligned}
\Pr(\rho > 1 + \varepsilon) &= \Phi\Big(\frac{\mu - \log(1 + \varepsilon)}{\sigma}\Big), \\
\Pr(\rho < 1 - \varepsilon) + \Pr(\rho > c) &= \Phi\Big(\frac{\log(1 - \varepsilon) - \mu}{\sigma}\Big) + \Phi\Big(\frac{\mu - \log c}{\sigma}\Big),
\end{aligned}
\end{equation}
with $\sigma = \sqrt{-2\mu}$. With $\mu = -\sigma^2/2$ the first summand of the second share has the argument $\sigma/2 + \log(1 - \varepsilon)/\sigma$, which increases with $\sigma$, and the second summand has the argument $-\sigma/2 - \log c/\sigma$, which increases up to $\sigma = \sqrt{2\log c} \approx 1.48$. The share of dropped tokens with negative advantage therefore grows until the mean lag reaches about $-1.1$ nats at $c = 3$. The argument of the first share is $-\sigma/2 - \log(1 + \varepsilon)/\sigma$, which increases up to $\sigma = \sqrt{2\log(1 + \varepsilon)} \approx 0.60$, a mean lag of about $-0.18$ nats at $\varepsilon = 0.2$.

\section{Reference Policy of PPO}\label{app:ppo}
PPO weights each token by the ratio $r_t = p_\theta(y_t\mid h_t)/\pi_\mathrm{ref}(y_t\mid h_t)$ to its reference, and its logarithm relates to the split of \cref{eq:split} as
\begin{equation}
\log r_t = \log p_\theta - \log \pi_\mathrm{ref} = \delta_t - \big(\log\pi_\mathrm{ref} - \log q\big).
\end{equation}
With the reference at the sampler's weights, $\pi_\mathrm{ref} = p_{\old}$ and $\log\pi_\mathrm{ref} - \log q = e_t$, so $\log r_t = \ell_t$ and the clip acts on the policy lag alone. With the reference at the trainer before the batch, $\pi_\mathrm{ref} = p_{\theta_\text{batch}}$ and
\begin{equation}
\log r_t = \ell_t - \big(\log p_{\theta_\text{batch}} - \log p_{\old}\big),
\end{equation}
which is the lag accumulated during the current batch alone. At the first of the $K$ updates per batch $\theta = \theta_\text{batch}$, so $r_t = 1$, the clip is inactive and the per-token gradient is $A_t\,s_t$, the policy gradient. The later updates move $r_t$ only by the change within the batch, so the clip bounds this change and leaves both the policy lag and the numerical discrepancy outside the ratio.

\section{Properties of DLC}\label{app:slc}
This section derives the behavior of the DLC update of \cref{eq:slc} at a single prefix, with $q_v = p_{\old}(v)\,e^{-e_v}$ and $\rho_v = p_\theta(v)/p_{\old}(v)$, so that $q_v\,\rho_v = p_\theta(v)\,e^{-e_v}$.

At a prefix with a constant advantage $A$, the expectation of the DLC update under the sampler is
\begin{equation}
\Ex_q\big[A\,(m\,\rho\,s - C)\big] = A\Big(\sum_v q_v\,m_v\,\rho_v\,s_v - \sum_{v\in H}\big(q_v\,m_v\,\rho_v - \kappa\,p_\theta(v)\big)\,s_v\Big)
= A\sum_{v\notin H}\big(q_v\,\rho_v - \kappa\,p_\theta(v)\big)\,s_v ,
\end{equation}
where the second step cancels the head and uses $\kappa\sum_{v\in H}p_\theta(v)\,s_v = -\kappa\sum_{v\notin H}p_\theta(v)\,s_v$ with the clip inactive on the tail. The remainder equals $A\sum_{v\notin H}p_\theta(v)\big(e^{-e_v} - \kappa\big)\,s_v$, which vanishes when the sampler's tail is $\kappa$ times the tail of $p_{\old}$. Up to this tail estimate DLC removes the drift of the discrepancy and of the lag, since the subtracted term equals the sampler's mean of the clipped score.

With $m_v = 1$ on the tokens of the prefix, the centering term is $C = \sum_v q_v\,\rho_v\,s_v$ up to the tail estimate, and the update at the prefix is
\begin{equation}
\Ex_q\big[a\,(\rho\,s - C)\big] = \sum_v q_v\,\rho_v\,a_v\,s_v - \bar a^q\sum_v q_v\,\rho_v\,s_v = \sum_v p_\theta(v)\,e^{-e_v}\,\big(a_v - \bar a^q\big)\,s_v .
\end{equation}
Subtracting the on-policy update $\sum_v p_\theta(v)\,(a_v - \bar a^p)\,s_v$ gives the token term of DLC,
\begin{equation}
\sum_v p_\theta(v)\big(e^{-e_v} - 1\big)\big(a_v - \bar a^p\big)\,s_v - \big(\bar a^q - \bar a^p\big)\sum_v p_\theta(v)\,e^{-e_v}\,s_v ,
\end{equation}
which has the form of \cref{eq:app-token} with the numerical discrepancy $e$ in place of the sampled-token error $\delta$ in the first term. The first term depends on $e$ alone, which stays fixed between synchronizations by \cref{app:parts}. The lag enters the second term through $\bar a^q - \bar a^p = \sum_v p_\theta(v)\big(e^{-\ell_v - e_v} - 1\big)\,a_v$, and there it multiplies $\sum_v p_\theta(v)\,e^{-e_v}\,s_v = \sum_v p_\theta(v)\big(e^{-e_v} - 1\big)\,s_v$, which vanishes with the numerical discrepancy. The lag therefore enters the token term of DLC at second order through its product with $e$, and under a stale sampler with $e = 0$ the token term vanishes exactly. To first order the token term of DLC is $-\Ex_p\big[(a - \bar a^p)\,e\,s\big]$, which keeps a fixed size and does not enter the loop of \cref{eq:loop}.

For a fixed sign of the advantage, the tokens with $m_v = 0$ change the update at the prefix by
\begin{equation}
-\sum_v q_v\,(1 - m_v)\,\rho_v\,\big(a_v - \bar a^q\big)\,s_v ,
\end{equation}
since both the weighted score and the centering term lose these tokens. The dropped tokens therefore return their lag-weighted share of the token term, as the truncated tokens of importance sampling do in \cref{eq:app-trunc}. Their share grows with the spread of the lag by \cref{app:ess}, so the clip has a validity range of the same kind as the truncation of importance sampling, and DLC resynchronizes the sampler where this range ends.

With $\theta = \old$ the lag ratio is one on each token, so $m_v = 1$ and $C_t = \sum_{v\in H}(q_v - \kappa\,p_\theta(v))\,s_v = \hat s_t$, the head estimate of score centering. With $q = p_{\old}$ the tail ratio is $\kappa = \big(1 - \sum_H p_{\old}\big)/\big(1 - \sum_H p_{\old}\big) = 1$ and $q_v\,\rho_v = p_\theta(v)$, so $C_t = \sum_{v\in H}p_\theta(v)\,(m_v - 1)\,s_v$, which vanishes with the clip inactive and leaves PPO with the sampler's weights as reference. The resynchronization sets $\old = \theta$, so $\rho_t = 1$, and $\ell_t$ and the lag part of $S_t$ vanish together, and the version gap $\Delta_{\old}$ restarts from zero in \cref{eq:app-unroll}. The numerical discrepancy is unchanged by the copy, so DLC keeps centering it after each resynchronization.

\section{Loss of DLC}\label{app:split}
The loss adds the centering term $C_t$ of \cref{eq:slc} to the PPO loss $L_\mathrm{PPO}$ with reference $\old$ in verl, with $\mathrm{sg}$ for the stop-gradient operator,
\begin{equation}
L_\mathrm{DLC} = L_\mathrm{PPO} + \sum_t A_t \sum_{v\in H}\mathrm{sg}\big[q_v\,m_v\,\rho_v - \kappa\,p_\theta(v)\big]\log p_\theta(v\mid h_t).
\end{equation}
The gradient of $L_\mathrm{PPO}$ is $-\sum_t m_t\,\rho_t\,A_t\,s_t$, and the gradient of the added term is $\sum_t A_t C_t$, so $-\nabla L_\mathrm{DLC} = \hat g_\mathrm{DLC}$. The indicator $m_v$ follows the sign of $A_t$ as $m_t$ does, so the loss computes the head term for both signs and keeps the one that matches the sign of the token's advantage. Both terms are averaged over the tokens of the batch as the PPO loss is. The head log-probabilities of $p_{\old}$ come from the forward pass at $\old$ that also gives $\ell_t$, and $\rho_v$ is clamped to the interval from $e^{-20}$ to $e^{20}$, as verl clamps the PPO ratio.

\end{document}